\documentclass{article}

\usepackage{microtype}
\usepackage{graphicx}
\usepackage{subcaption}
\usepackage{booktabs} 
\usepackage{cuted}

\usepackage{hyperref}

\usepackage[accepted]{icml2026}

\usepackage{amsmath}
\usepackage{amssymb}
\usepackage{mathtools}
\usepackage{amsthm}

\usepackage{algorithm}
\usepackage{algorithmic}

\usepackage[table]{xcolor}
\usepackage{makecell}
\usepackage{tabularx}
\usepackage{multirow}
\usepackage{wrapfig}
\usepackage{caption}
\usepackage{pifont}
\newcommand{\cmark}{\textcolor{green}{\ding{51}}} 
\newcommand{\xmark}{\textcolor{red}{\ding{55}}}   

\definecolor{GroupGray}{HTML}{F8EEF2}
\definecolor{GroupBlue}{HTML}{EFF6FC}
\definecolor{GroupBeige}{HTML}{FDF6E3}
\definecolor{ForestGreen}{HTML}{228B22}

\usepackage[capitalize,noabbrev]{cleveref}

\icmltitlerunning{Temporal Understanding for Agentic Autonomous Driving via VLMs}

\begin{document}

\twocolumn[
  \icmltitle{From Segments to Scenes: Temporal Understanding for Agentic
   Autonomous Driving via Vision-Language Models}

    \icmlsetsymbol{equal}{*}
    \icmlsetsymbol{lead}{\textdagger}

  \begin{icmlauthorlist}
    \icmlauthor{Kevin Cannons}{equal,inst1}
    \icmlauthor{Saeed Ranjbar Alvar}{equal,inst1}
    \icmlauthor{Mohammad Asiful Hossain}{inst1}
    \icmlauthor{Ahmad Rezaei}{inst1}
    \icmlauthor{Mohsen Gholami}{inst1}
    \icmlauthor{Alireza Heidarikhazaei}{inst1}    
    \icmlauthor{Zhou Weimin}{inst2}   
    \icmlauthor{Yong Zhang}{lead,inst1}   
    \icmlauthor{Mohammad Akbari}{lead,inst1}     
  \end{icmlauthorlist}

  \icmlaffiliation{inst1}{Huawei Technologies Canada Co., Ltd.}
  \icmlaffiliation{inst2}{Huawei Cloud}

  \icmlcorrespondingauthor{Kevin Cannons}{kevin.cannons@huawei.com}
  \icmlcorrespondingauthor{Saeed Ranjbar Alvar}{saeed.ranjbar.alvar1@huawei.com}

  \icmlkeywords{Autonomous Driving, Agentic AI, Temporal Understanding, Vision-Language Models, Benchmarking}

    \vskip 0.05in
    \centering
    \href{https://github.com/vbdi/tad_bench}{\includegraphics[height=1.25em]{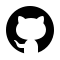} Code} \quad
    \href{https://huggingface.co/datasets/vbdai/TAD}{\includegraphics[height=1.25em]{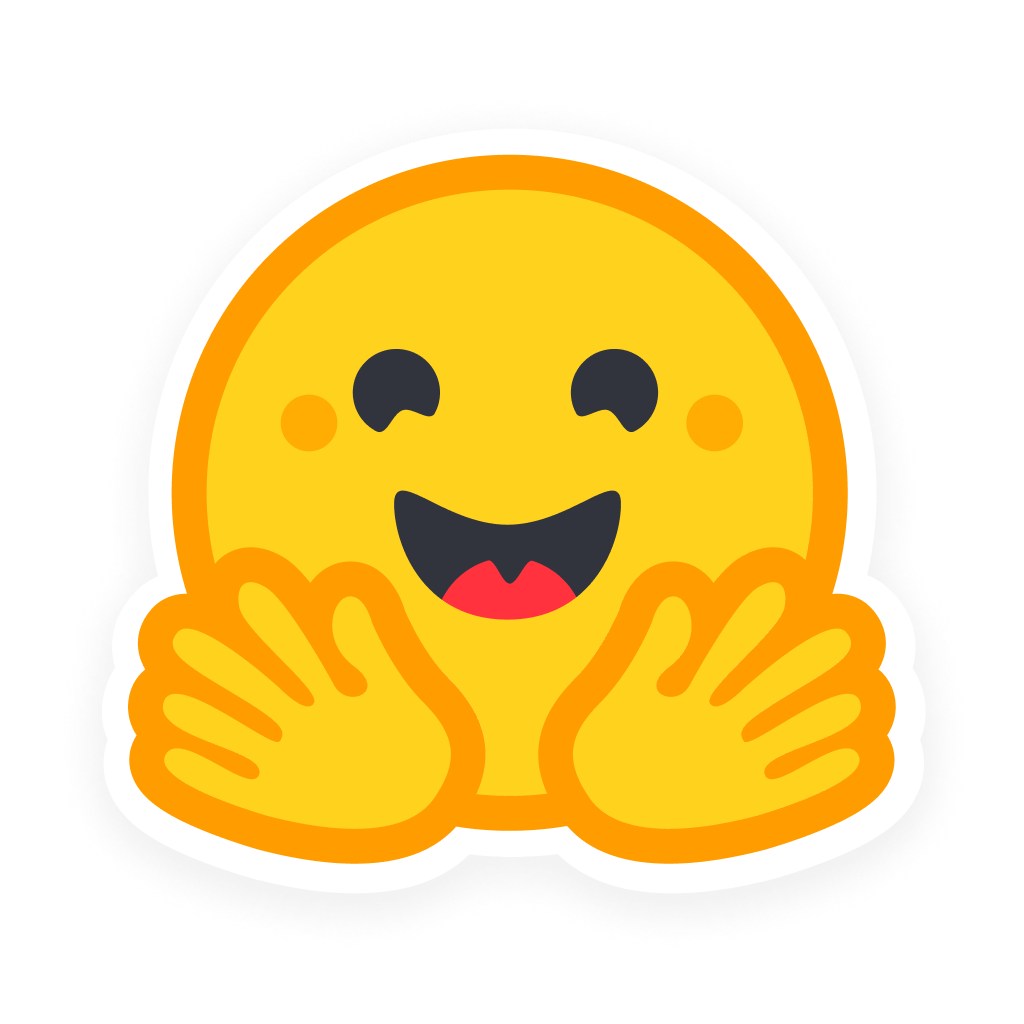} Dataset}
    \vspace{-.1in}
]

\printAffiliationsAndNotice{%
  \textsuperscript{*} Equal contribution. \textsuperscript{\textdagger} Co-leads.
}

\begin{strip}    
  \centering
  \includegraphics[width=0.62\textwidth]{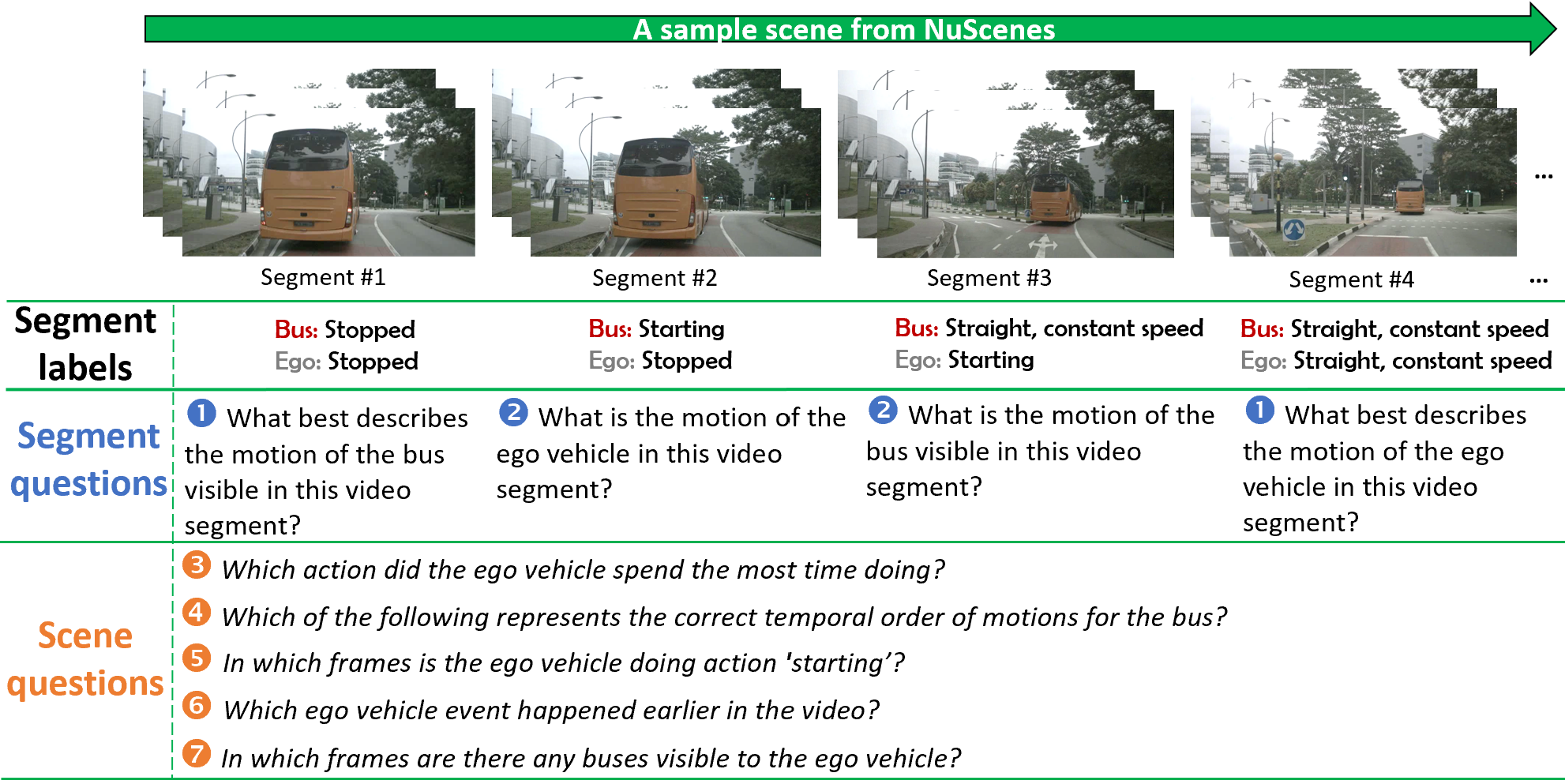}\hfill
  \includegraphics[width=0.36\textwidth]{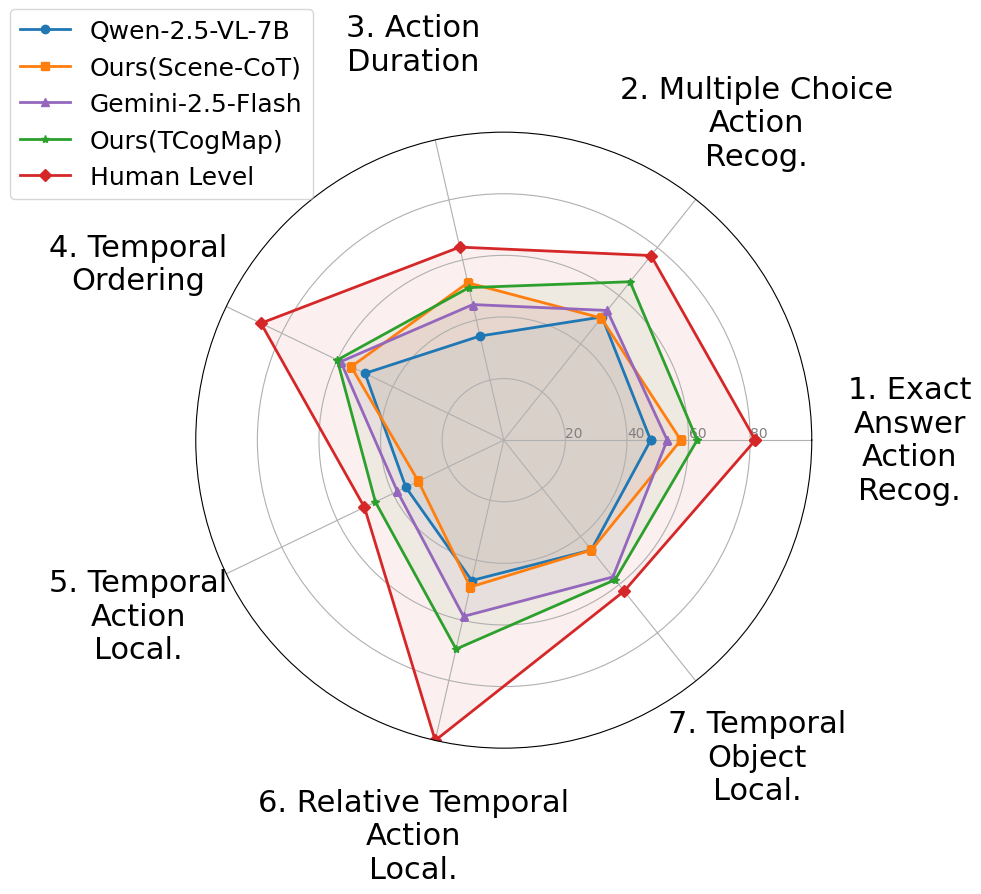}
  \captionof{figure}{Overview of the proposed \textbf{TAD} benchmark. \textbf{Left}: An example of labels and question types. Each segment covers approximately 5 seconds of a video. Two question types pertain to actions within video segments, while the remaining five require information from the entire video. \textbf{Right:} Performance of open-source VLMs, the proposed training-free solutions (\textbf{Scene-CoT} and \textbf{TCogMap}), a closed-source VLM, and human performance across the 7 benchmark tasks in \textbf{TAD}.}
  \label{fig:teaser_alt}
\end{strip}

\begin{abstract}
Vision-Language Models (VLMs) are increasingly deployed as the perception and reasoning backbone of autonomous agents acting in the wild, with autonomous driving (AD) being one of the most safety-critical instances. Reliable temporal understanding is essential for such agents to anticipate events, attribute causes, and act safely in dynamic environments, yet this remains a significant challenge even for state-of-the-art (SoTA) VLMs. Prior video benchmarks have emphasized other content (sports, cooking, etc.), yet no existing benchmark focuses \textit{exclusively} on temporal understanding for \textit{both short- and long-form} AD footage. To fill this gap, we present the Temporal Understanding in Autonomous Driving (TAD) benchmark, comprising nearly 6{,}000 question-answer (QA) pairs across 7 tasks, and evaluate 9 closed- and open-source generalist as well as AD-specialist models. Current SoTA models perform substantially below human accuracy on TAD. To improve the temporal reasoning of VLM-based driving agents, we propose two novel \textit{training-free} solutions: \textbf{Scene-CoT}, which uses Chain-of-Thought (CoT) reasoning, and \textbf{TCogMap}, which incorporates an ego-centric temporal cognitive map produced by a trajectory-analysis module that operates as an agentic tool around the VLM. Integrated with existing VLMs, our methods improve average accuracy on TAD by up to 17.72\% and by up to 10.35\% on STSBench. By introducing TAD, benchmarking SoTA models, and proposing effective enhancements, this work aims to catalyze further progress on temporal understanding for agentic AD systems operating in the wild. The benchmark and evaluation code are available at ${\href{https://huggingface.co/datasets/vbdai/TAD}{\text{Hugging Face}}}$ and ${\href{https://github.com/vbdi/tad_bench}{\text{GitHub}}}$, respectively.
\end{abstract}


\section{Introduction}
\label{sec:intro}
Vision-Language Models (VLMs) are rapidly becoming the perception-and-reasoning engine of autonomous agents that must act in open-ended real-world environments. Among such agents, autonomous-driving (AD) vehicles are arguably the most safety-critical: they must continuously perceive, reason, and act in cluttered, dynamic scenes where errors can be catastrophic. Reliable \textit{temporal} understanding of video is a prerequisite for any such agent --- to anticipate events, attribute causes, plan over multiple seconds, and produce trustworthy explanations of its own behavior. Accordingly, evaluating VLMs on video understanding tasks has emerged as an important research direction~\cite{MVBench,VideoMME,SEEDBench,TempCompass,STIBench,TemporalBench}. However, existing benchmarks have generally focused on specific types of video, such as sports~\cite{caba2015activitynet}, cooking~\cite{zhou2018weakly}, and movies~\cite{shu2025video}.

AD videos exhibit unique challenges compared to general video data:
\textbf{1) Temporal Scale.} Large variation in the temporal scale of activities, requiring methods that localize events of varying durations within videos of arbitrary length.
\textbf{2) Ego-Centric View.} The camera viewpoint makes understanding the ego-vehicle's actions challenging, since the ego is not observable in the frames.
\textbf{3) Fine-grained Actions.} Vehicle actions can be subtle and may require additional cues for robust detection (e.g., a gradual lane change vs.\ driving straight).
Currently, there is no \textit{AD-exclusive benchmark} for assessing VLMs on these unique aspects of \textit{temporal understanding} that includes \textit{both short and long videos}.

In light of this gap, this paper proposes the \textbf{\underline{T}}emporal Understanding in \textbf{\underline{A}}utonomous \textbf{\underline{D}}riving (TAD) benchmark, which evaluates a range of temporal-understanding tasks for VLM-based driving agents. TAD is built on the popular NuScenes dataset~\cite{nuscenes} and consists of 5{,}861 QA pairs across 150 videos. Most tasks focus on \textbf{``scene-level''} (i.e., \textbf{video-level}) understanding, but two consider \textbf{``segment-level''} action recognition (short video segments $\leq$ 5 seconds). Inclusion of both scene and segment questions is a novel characteristic of TAD. Segment questions assess fine-grained vehicle actions (e.g., left-hand turns vs.\ changing lanes to the left); scene questions require understanding the temporal dynamics among atomic actions that together form a longer video. In addition, 4{,}481 segment-level vehicle action annotations were created. \cref{fig:teaser_alt} shows an example.

The paper presents an extensive evaluation on 9 unique VLMs across 30 system configurations, including open-source generalist, AD-specialist, and closed-source models. The results in \cref{fig:teaser_alt} show a significant gap between VLMs and humans --- a gap that directly limits the reliability of any agentic driving system that uses such a VLM as a component. To address this gap, two novel solutions, \textbf{Scene-CoT} and \textbf{TCogMap}, are proposed. Scene-CoT leverages Chain-of-Thought (CoT) reasoning to answer AD questions by breaking down scene motions. TCogMap constructs an ego-centric temporal cognitive map from the ego vehicle's trajectory, used to enrich the VLM's input. Crucially, TCogMap operates as a lightweight \textit{agentic} wrapper around the VLM: a deterministic trajectory-analysis tool that produces structured motion summaries that the VLM consumes alongside the raw frames.

The major contributions of this work are:
\begin{itemize}
    \item Creation of TAD, the first QA benchmark \textit{exclusively for AD} that evaluates \textit{temporal understanding} for \textit{both short- and long-form videos}, providing a testbed relevant to agentic driving systems operating in the wild.
    \item A new dataset of fine-grained, human-annotated action labels for ego and non-ego vehicles across 150 NuScenes videos.
    \item A comprehensive evaluation of SoTA open-source, closed-source and specialist VLMs on TAD.
    \item Introduction of Scene-CoT and TCogMap, training-free methods that improve average performance on TAD by up to 17.72\% and on STSBench~\cite{stsbench} by up to 10.35\%.
\end{itemize}

\section{Related Work}
\label{sec:related_work}

\subsection{Temporal Understanding in General Videos}
The temporal understanding of general-purpose videos has been explored through various tasks and benchmarks. Key tasks include \textit{Temporal Localization} --- specifying when an event occurs~\cite{LITA,TimeMarker,NumberIt,GroundedVideoLLM} --- and \textit{Dense Video Captioning}, which localizes and describes every event~\cite{Krishna_2017,Zhou_2018,Deng_2021,Wang_2021,Yang_2023,jin2024tod3cap,Kim_2024,Zhou_2024}. Some benchmarks, like MVBench~\cite{MVBench} and Video-MME~\cite{VideoMME}, assess a combination of spatial and temporal abilities. Others, such as TempCompass~\cite{TempCompass} and TemporalBench~\cite{TemporalBench}, focus on fine-grained temporal reasoning across daily life, sports, and documentaries. None target the agentic driving setting, where temporal reasoning is a prerequisite for performing actions safely.

\subsection{Benchmarks and Methods for VLM-Based AD}
VLMs have emerged as a promising direction for enhancing AD systems, offering improved explainability and the ability to leverage multiple modalities~\cite{jin2024tod3cap,Marcu_2024_ECCV,Ma_2024_ECCV,Sima_2024_ECCV,Nie_2024_ECCV,Ding_2024_CVPR,Chen_2024_ECCV,Pan_2024_CVPR,Shao_2024_CVPR,Tian_2025_CVPR,Wang_2025_CVPR,Xu_2025_CVPR}. Their applications range from conversational driving assistants~\cite{Ma_2024_ECCV} and VQA systems~\cite{Marcu_2024_ECCV,Sima_2024_ECCV} to integration within the driving pipeline itself~\cite{xu2024drivegpt4,Pan_2024_CVPR,Shao_2024_CVPR,Chen_2024_ECCV,Tian_2025_CVPR,Wang_2025_CVPR,Xu_2025_CVPR,wang2025embodied}. In all such settings, the VLM acts as one component of a larger agentic system, and its temporal-reasoning failure modes propagate directly to that agent's behavior.

Several VLM-oriented benchmarks have been released in recent years to assess different AD-related tasks. LingoQA~\cite{Marcu_2024_ECCV}, DriveLM~\cite{Sima_2024_ECCV}, and VLADBench~\cite{VLADBench} offer broad QA datasets spanning perception, prediction, and planning. Other works focus on narrower aspects: BDD-X~\cite{BDDX} provides ego-action descriptions and reasons; NuScenes-QA~\cite{NuScenesQA} and NuPrompt~\cite{NuPrompt} are perception-oriented; DRAMA~\cite{DRAMA} and Rank2Tell~\cite{Rank2Tell} annotate critical scene objects with action suggestions; Ego3D-Bench~\cite{Ego3D} considers 3D spatial understanding; SURDS~\cite{SURDS} targets fine-grained spatial understanding; CODA-LM~\cite{Chen_2025_WACV} evaluates corner cases; and MetaVQA~\cite{wang2025embodied} provides spatial and embodiment-related questions for closed-loop simulation.

The most closely-related works to TAD are ELM~\cite{zhou2024embodied}, STSBench~\cite{stsbench}, and STIBench~\cite{STIBench}. ELM is a training-oriented dataset; only two of its AD tasks require knowledge of previous frames, and its videos are 3.5\,s long. STSBench contains only event/action recognition for short ($\leq$3\,s) segments. STIBench uses 20\,s videos for AD but focuses on general embodied scene understanding (e.g., camera-pose estimation), and most of its data is indoor.

In contrast, TAD is the first benchmark to exclusively assess temporal understanding in AD via both segment- and scene-level questions. \cref{tab:benchmark_comparisons} summarizes the most closely-related QA benchmarks; TAD is the second-largest video-only benchmark focused on temporal understanding. Note that ELM's \#QAs is not directly comparable: it is training-oriented, mixes AD with other ego-centric imagery, and varies the input modality by question type.

\begin{table}[t]
\centering
\caption{AD benchmarks for spatiotemporal understanding. ``\#QAs'': number of questions. ``I/V'': image or video. ``Seg./Scene Qs'': segment- and scene-level questions.}
\label{tab:benchmark_comparisons}
\footnotesize
\setlength{\tabcolsep}{4pt}
\begin{tabular}{l r ccc}
\toprule
\textbf{Benchmark} & \textbf{\#QAs} & \textbf{I/V} & \textbf{Seg.} & \textbf{Scene} \\
\midrule
DriveLM~\cite{Sima_2024_ECCV}      & 15k    & I & \xmark & \xmark  \\
DriveBench~\cite{Xie_2025_ICCV}    & 20k    & I & \xmark & \xmark  \\
SURDS~\cite{SURDS}                 & 9.2k   & I & \xmark & \xmark  \\
Ego3D~\cite{Ego3D}           & 8.6k   & I & \xmark & \xmark  \\
DriveLLM-o1~\cite{ishaq2025drivelmm}& 4.6k  & I & \xmark & \xmark  \\
ELM~\cite{zhou2024embodied}        & 1525k  & I\,\&\,V & \cmark & \xmark  \\
LingoQA~\cite{Marcu_2024_ECCV}     & 1k     & V & \cmark & \xmark  \\
VLADBench~\cite{VLADBench}         & 12k    & V & \xmark & \cmark  \\
STSBench~\cite{stsbench}           & 971    & V & \cmark & \xmark  \\
STIBench~\cite{STIBench}           & 2k     & V & \xmark & \cmark  \\
\midrule
\textbf{TAD (Ours)}                & 5.8k   & V & \cmark & \cmark  \\
\bottomrule
\end{tabular}
\end{table}

\section{TAD Benchmark}
\label{sec:tad-benchmark}

\subsection{Benchmark Annotations}
The TAD benchmark is built on top of the NuScenes~\cite{nuscenes} validation split, which contains 150 videos, each roughly 20 seconds in length, with accompanying sensor data and 3D bounding-box annotations. TAD focuses on the following fine-grained vehicle classes defined in NuScenes: car, bus, bicycle, construction vehicle, motorcycle, trailer, and truck. All other object annotations were ignored.

\begin{figure*}[t]
\centering
  \includegraphics[width=0.95\textwidth]{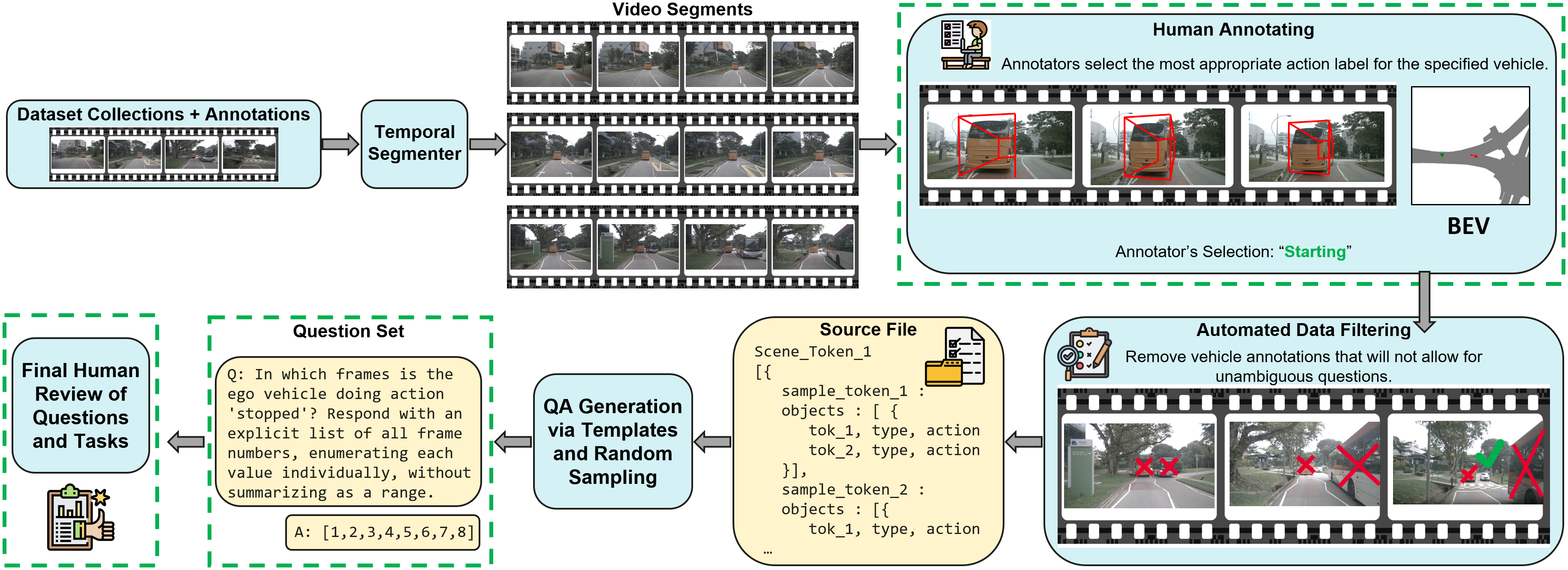}
 \caption{TAD annotation and question generation pipeline. Boxes with green dash: steps with human verification and quality assessment.}
 \label{fig:annotation_pipeline}
 \end{figure*}

NuScenes does not provide action labels for scene objects, yet action understanding is a critical aspect of understanding the temporal dynamics of a scene, and thus, a critical aspect of the situational awareness needed by an AD agent. Hence, the first annotation task to create TAD was to label vehicle actions in NuScenes, which required a pre-processing step that divides each video into segments, as shown in \cref{fig:annotation_pipeline}. For vehicle action understanding, employing a short segment-level approach is appropriate, since an action typically cannot be discerned from a single frame, and in a long video, multiple actions may occur. For TAD, five-second segments were selected, since most vehicle atomic actions are completed within five seconds. Thus, each NuScenes video was divided into ten uniformly-distributed, overlapping segments.

To streamline annotation and focus on the most relevant vehicles, each segment was filtered to include only vehicles within 50\,m of the ego. This threshold was empirically set to retain the most relevant vehicles while optimizing human annotation effort. Vehicles with negligible trajectory displacement were automatically labeled as \textit{stopped} and excluded from manual labeling.

Video segments with the corresponding bird's-eye view (BEV) visualization, including the trajectories of the ego and nearby vehicles, were provided to annotators. Annotators were asked to label a vehicle's action corresponding to the dominant behavior during the video segment. Accordingly, only one action label was assigned to a vehicle per segment. The eight action categories used for TAD are: 1) traveling straight at constant speed, 2) stopping, 3) stopped, 4) starting, 5) turning left, 6) turning right, 7) changing lanes to the left, and 8) changing lanes to the right. In addition to the TAD benchmark, the vehicle-action annotations are released as a standalone resource that can be used to evaluate vehicle action recognition.

The final step of the annotation pipeline generates a JSON-formatted source file that stores the object and action labels per segment. These files constituted the raw materials used to construct the TAD QA set.

\subsection{QA Creation}
As shown in \cref{fig:annotation_pipeline}, QA generation followed a template-driven approach. Seven QA templates were applied to the NuScenes videos. Logical checks were also included to verify that each question was unambiguous, e.g., ensuring that the objects referenced in each question are uniquely identifiable in the imagery. A final human-verification stage was included to further improve the quality of the QA pairs and task definitions.

After running the QA generation pipeline, 5{,}861 QAs were created for TAD, distributed across seven tasks. Several tasks --- \textit{Temporal Ordering}, \textit{Temporal Action Localization}, \textit{Relative Temporal Action Localization}, and \textit{Temporal Object Localization} --- are designed to challenge VLMs on \textbf{when} specific actions or events occur. \textit{Temporal Ordering} and \textit{Relative Temporal Action Localization} require the VLMs to understand the \textbf{temporal ordering} of events. The \textit{Action Duration} task does not focus on when events happen, but \textbf{how long} they last. The final two tasks, \textit{Exact Answer Action Recognition} and \textit{Multiple Choice Action Recognition}, consider the VLM's ability to understand \textbf{fine-grained actions} in temporally cropped video segments. Sample questions are shown in \cref{fig:teaser_alt}; detailed descriptions of each question type are provided in the supplementary materials.

\subsection{TAD Benchmark Statistics}
\textbf{Question Distribution.} The question distribution is shown in \cref{tbl:ego_non_ego_count}. In addition to the ``Total'' column, a breakdown is provided showing the number of ego vs.\ non-ego questions. All question types are well-represented, with a minimum of 80 per category. Two question categories, \textit{Action Duration} and \textit{Relative Temporal Action Localization}, only consider the ego vehicle. Conversely, \textit{Temporal Object Localization} only includes questions about non-ego car objects, since this question considers when those objects are visible to the ego. The overall split between ego and non-ego vehicle questions is almost even (53.7\% and 46.3\%, respectively), so there is no significant benchmark bias favouring ego or non-ego objects.

\textbf{Object Distribution.}
A breakdown of the vehicle types considered in TAD, as well as their counts, is shown in \cref{tab:object_count}. The benchmark considers all vehicle categories relatively evenly, the lone exception being ``Emergency Vehicle'', which is rare in the original data itself. Note that some questions refer to ``Non-Ego'' vehicles, which corresponds to any vehicle other than the ego. Further stats, including instance count per action category, are in the supplementary material.

\begin{table}[!t]
\centering
\caption{Question counts per task in TAD. \textbf{AR}: Action Recognition. \textbf{AL}: Action Localization. \textbf{OL}: Object Localization.}
\begin{tabular}{lccc}
\toprule
\textbf{Category} & \textbf{Ego} & \textbf{Non-Ego} & \textbf{Total} \\
\midrule
Exact Answer AR & 1500 & 1104 & 2604 \\
Multiple Choice AR & 1030 & 756 & 1786 \\
Action Duration & 124 & 0 & 124 \\
Temporal Ordering & 62 & 18 & 80 \\
Temporal AL & 338 & 432 & 770 \\
Relative Temporal AL & 92 & 0 & 92 \\
Temporal OL & 0 & 405 & 405 \\
\midrule
\textbf{Total Dataset} & \textbf{3146} & \textbf{2715} & \textbf{5861} \\
\bottomrule
\end{tabular}
\label{tbl:ego_non_ego_count}
\end{table}

\section{Proposed Methods}

We propose two training-free, plug-in methods that wrap any VLM to improve its temporal understanding for AD agents: \textbf{Scene-CoT} and \textbf{TCogMap}. Both adopt the now-standard \textit{tool-augmented} pattern of agentic systems, a structured pre-processing step that decomposes the video and produces text that the VLM consumes alongside the raw frames.

\subsection{Scene-CoT: CoT for Temporal Understanding}
The proposed Scene-CoT framework introduces a CoT-based~\cite{cot-llm} approach for structured scene reasoning in dynamic driving environments, consisting of three modules: 1) Video partitioning, 2) CoT reasoning, and 3) Large Language Model (LLM) QA. 
The system modules are described next, while a figure showing the Scene-CoT architecture is provided in the supplementary materials.

\textbf{Video Partitioning.}
Scene-CoT first partitions the input video into smaller segments. Let $\mathcal{V}$ denote the original video and $S=\{s_1, s_2, \dots, s_l\}$ be the set of $l$ segments. The segment length and overlap are configurable to meet design and resource constraints. In the current implementation, segments are uniformly distributed across $\mathcal{V}$, each lasting 5 seconds with approximately $50\%$ overlap between consecutive segments. This choice is suitable for driving scenarios, as a five-second window with overlap provides sufficient temporal context to capture common maneuvers (e.g., lane changes, turns) while maintaining responsiveness to rapidly changing scene dynamics. Exploration of more sophisticated segmentation strategies is left for future work.

To reduce pixel-wise redundancy from frame to frame and to limit the number of visual tokens input to the VLM, only a subset of frames is used to form the CoT descriptions. Specifically, four frames are uniformly sampled across each segment, i.e., $\{f_j^1, f_j^2, f_j^3, f_j^4\}$, where $f$ denotes a frame and $j$ corresponds to a segment number.

\textbf{CoT Reasoning.}
For each segment, $s_j$, the VLM is prompted to generate a reasoning trace sufficient to answer the questions in TAD. To that end, a four-step CoT reasoning procedure is designed: \textbf{1) Scene description.} The VLM provides a high-level description of the scene. \textbf{2) Ego vehicle motion.} The VLM focuses on the ego car motion and estimates its action. \textbf{3) Motion of nearby vehicles.} The VLM isolates other vehicles close to the ego and describes their motion. \textbf{4) Summary formatting.} The VLM generates a final JSON-style summary containing fields for motions of all vehicles and nearby-vehicle identifiers.
The final output for segment $s_j$, denoted $c_j$, is the concatenation of the scene description (Step 1) with the summary (Step 4).

\textbf{LLM-Based QA.}
The LLM takes as input the benchmark question, $q$, together with the temporally ordered CoT reasoning outputs for segments $\mathcal{C} = (c_1, c_2, c_3, \ldots, c_l)$. Since the first step of Scene-CoT divides the video into short, temporally ordered segments, the resulting CoT outputs support questions that require temporal reasoning and ordering. Formally, the final answer is
\begin{equation}
a_{\textnormal{Scene-CoT}} = \mathcal{L}(\mathcal{C}, q),
\end{equation}
where $\mathcal{L}$ denotes the LLM. Additional details regarding Scene-CoT are provided in the supplementary materials.

\begin{table}[!t]
\centering
\caption{Statistics of vehicle types in the TAD questions.}
\renewcommand{\arraystretch}{0.99}
\begin{tabular}{lr@{\hskip 12pt}lr}
\toprule
\textbf{Type} & \textbf{Count} & \textbf{Type} & \textbf{Count} \\
\midrule
Bicycle & 153 & Motorcycle & 257 \\
Bus & 286 & Non-Ego & 432 \\
Car & 749 & Trailer & 108 \\
Construction Vehicle & 123 & Truck & 594 \\
Emergency Vehicle & 13 & \textbf{Total} & \textbf{2715} \\
\bottomrule
\end{tabular}
\label{tab:object_count}
\end{table}

\subsection{TCogMap: Ego-Centric Temporal Cognitive Maps}

\begin{figure*}[t]
 \centering
  \includegraphics[width=0.95\textwidth]{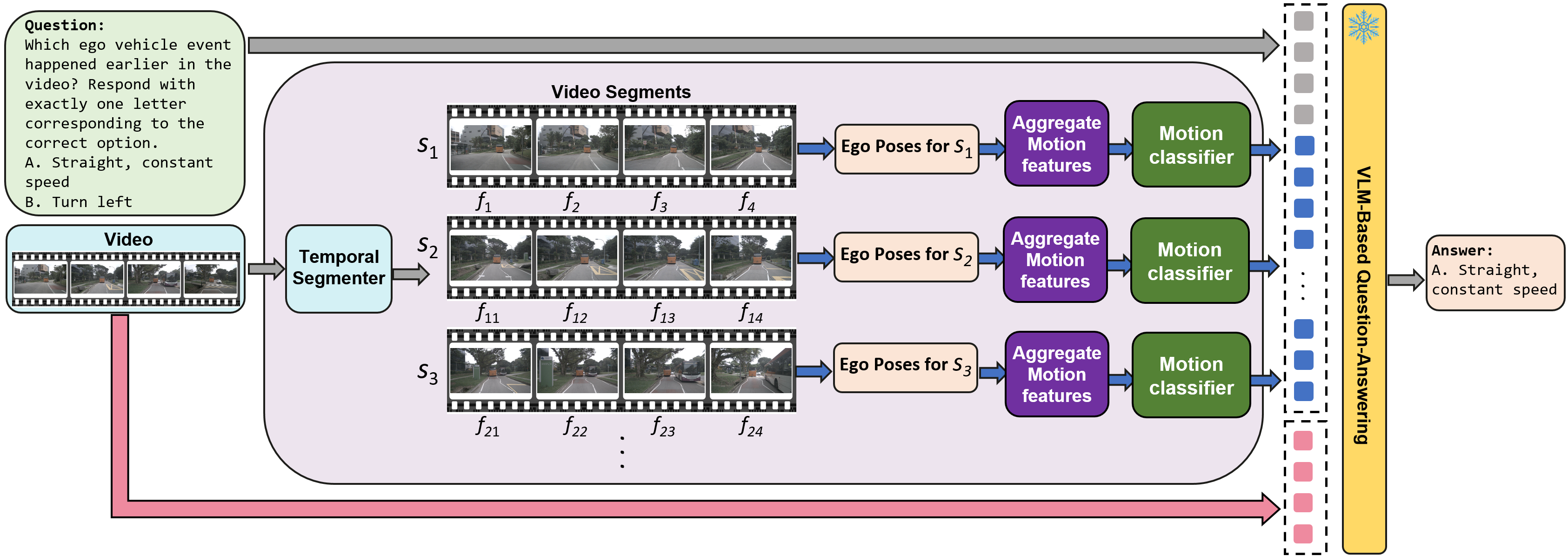}
 \caption{Overview of the proposed TCogMap method. A deterministic trajectory-analysis tool produces structured ego-motion summaries that augment the VLM's input --- a lightweight agentic pattern around the VLM.}
 \label{fig:TCogMap}
 \end{figure*}

The second proposed method, TCogMap, is a solution tailored to temporal understanding in the AD domain, as it leverages a vehicle trajectory-analysis module that operates as an agentic tool around the VLM. \cref{fig:TCogMap} shows the three steps in TCogMap, described next.

\textbf{Video Partitioning.}
TCogMap uses the same video partitioning module as Scene-CoT, resulting in a set of video segments, $S = \{s_1, s_2, s_3, \dots, s_l\}$.

\textbf{Ego-Vehicle Temporal Cognitive Map.}
Recent works have shown that generating a so-called ``cognitive map'' representation assists a VLM in question-answering~\cite{Ego3D,CognitiveMap}. Building on this insight, the proposed approach extends cognitive maps to the video domain for improved temporal understanding in outdoor and dynamic AD scenes. A full extension that includes information about all objects is appealing, but may overload the VLM with extraneous information. TCogMap therefore considers only the most important object in a driving video: the ego vehicle.

\newcommand{\Vstat}{V_{\text{stat}}}
\newcommand{\Vstop}{V_{\text{stopping}}}
\newcommand{\Pturn}{\Psi_{\text{turn}}}
\newcommand{\Vylc}{V_{y, \text{lc}}}
\newcommand{\Vxlc}{V_{x, \text{lc}}}

\begin{algorithm}[t]
\caption{Ego-Vehicle Motion Classification}
\label{alg:motion-classification-concise}
\footnotesize
\begin{algorithmic}
\STATE {\bfseries Input:} Ego-poses $P = \{p_0, \dots, p_{N-1}\}$, each $p_i$ with translation $\mathbf{x}_i$, quaternion $\mathbf{q}_i$, timestamp $t_i$.
\STATE {\bfseries Input:} Thresholds $\Vstat, \Vstop, \Pturn, \Vylc, \Vxlc$.
\STATE {\bfseries Output:} Motion label (e.g., ``Turn left'').
\FOR{$i=1$ {\bfseries to} $N-1$}
    \STATE $\mathbf{v}_i \leftarrow (\mathbf{x}_i - \mathbf{x}_{i-1}) / (t_i - t_{i-1})$
    \STATE $\omega_i \leftarrow \lVert \mathbf{v}_i[0\!:\!2] \rVert$
\ENDFOR
\IF{$(\sum_{i=1}^{N-1} [\omega_i < \Vstat])/(N-1) > 0.5$}
    \STATE {\bfseries return} ``Stopped''
\ENDIF
\STATE $\Delta\psi \leftarrow \text{yaw}(\mathbf{q}_{N-1}) - \text{yaw}(\mathbf{q}_0)$
\STATE $\omega_{\text{start}} \leftarrow \omega_1$,\ \ $\omega_{\text{end}} \leftarrow \omega_{N-1}$
\STATE $\{\mathbf{v}_{\text{local}, i}\} \leftarrow \text{ToLocalFrame}(\{\mathbf{v}_i\}, \{\mathbf{q}_i\})$
\STATE $\bar{v}_{x}, \bar{v}_{y} \leftarrow \text{mean}(\{\mathbf{v}_{\text{local}, i}\})$
\IF{$|\Delta\psi| > \Pturn$}
    \STATE {\bfseries return} ``Turn left'' if $\Delta\psi > 0$, else ``Turn right''
\ELSIF{$|\bar{v}_{y}| > \Vylc$ \AND $|\bar{v}_{x}| > \Vxlc$}
    \STATE {\bfseries return} ``Change lane left'' if $\bar{v}_{y} > 0$, else ``Change lane right''
\ELSIF{$\omega_{\text{start}} < \Vstop$ \AND $\omega_{\text{end}} > 1.5\Vstop$}
    \STATE {\bfseries return} ``Starting''
\ELSIF{$\omega_{\text{start}} > 1.5\Vstop$ \AND $\omega_{\text{end}} < \Vstop$}
    \STATE {\bfseries return} ``Stopping''
\ELSE
    \STATE {\bfseries return} ``Straight''
\ENDIF
\end{algorithmic}
\end{algorithm}

TCogMap analyzes the sequence of ego-vehicle poses (translation, rotation, and timestamp, all provided with NuScenes) to classify the ego's motion per segment. The procedure, detailed in \cref{alg:motion-classification-concise}, computes frame-wise global velocities and 2D speeds via finite differences (L4-7). A preliminary check classifies the segment as ``Stopped'' if instantaneous speeds fall below a stationary threshold, $\Vstat$ (L8-10). For segments with significant motion, dynamic features are extracted: the change in yaw $\Delta\psi$, initial and final speeds, and the average lateral/forward velocities $\bar{v}_{y}$, $\bar{v}_{x}$. Each global velocity is transformed into the ego's local frame using its rotation $\mathbf{R}_i$:
\begin{equation}
\mathbf{v}_{\text{local}, i} = \mathbf{R}_i^\top \mathbf{v}_i,
\end{equation}
and averages are computed over the segment. A hierarchical rule-based classifier (L14--24) then assigns one of eight motion labels, translating complex kinematics into a concise summary that the VLM can consume directly --- a deterministic ``tool'' that produces structured context for the agent's reasoning step.

\textbf{VLM-Based QA.}
The final step is to answer the question. The VLM, $\phi$, is provided with all $K$ frames of the input video, $\mathcal{F} = \{F_k\}_{k=1}^K$, the question $q$, and the structured ego-motion summary $\mathcal{M} = \{m_i\}_{i=1}^l$. To provide clear temporal context, each $m_i$ is formatted as a string specifying the corresponding frames (e.g., ``Motion summary for Frame1 to Frame7: The ego-vehicle is stopped''). The final answer is then
\begin{equation}
a_{\textnormal{TCogMap}} = \phi(\mathcal{F}, \mathcal{M}, q).
\end{equation}

\section{TAD Benchmarking}

\setlength{\tabcolsep}{4pt}
\begin{table*}[t!]
    \caption{Main results for open- and closed-source generalist VLMs as well as AD-specific VLMs. Bold and underline indicate best and second-best results when comparing the four system variants for each open-source VLM. +EP: raw textual ego pose information is included in the VLM's input context. \colorbox{GroupBlue}{Gen}: generalist; \colorbox{GroupBeige}{Drv}: driving specialist. \textbf{EA}: exact answer, \textbf{MC}: multiple choice.}
    \centering
    \begin{tabular}{l|c|ccccccc|c}
        \toprule
        \textbf{Method}
        & \textbf{Type}
        & \makecell{\textbf{EA}\\\textbf{Act.}\\\textbf{Recog.}}
        & \makecell{\textbf{MC}\\\textbf{Act.}\\\textbf{Recog.}}
        & \makecell{\textbf{Action}\\\textbf{Dur.}}
        & \makecell{\textbf{Temp.}\\\textbf{Ord.}}
        & \makecell{\textbf{Temp.}\\\textbf{Act.}\\\textbf{Local.}}
        & \makecell{\textbf{Rel.}\\\textbf{Temp.}\\\textbf{Local.}}
        & \makecell{\textbf{Temp.}\\\textbf{Obj.}\\\textbf{Local.}}
        & \textbf{Avg} \\ \midrule
        Human level &  & 81.68 & 76.67 & 64.29 & 87.50 & 50.21 & 100 & 62.69 & 74.72 \\
        Chance level &  & 12.17 & 24.24 & 25.81 & 30.00 & 41.70 & 47.83 & 58.85 & 34.37 \\ \midrule \midrule
    \rowcolor{GroupBlue} GPT-5-mini~\cite{gpt5} & Gen & 54.57 & 52.86 & 45.97 & 70.00 & 26.79 & 73.91 & 40.18 & 52.04 \\
    \rowcolor{GroupBlue} Gemini-2.5-Flash~\cite{gemini} & Gen & 53.07 & 53.86 & 45.16 & 58.75 & 38.35 & 58.70 & 56.81 & 52.10  \\ \midrule \midrule
    \rowcolor{GroupBlue} \textit{Qwen2.5-VL-7B}~\cite{qwenvl} & Gen & 47.77 & \underline{51.18} & 34.68 & 50.00 & \underline{35.28} & 46.74 & 45.60 & 44.46 \\
    \rowcolor{GroupBlue} \textit{Qwen2.5-VL-7B+EP} & Gen & 46.54 & 49.50 & 35.48 & 48.75 & 35.12 & \underline{57.61} & 37.27 & 44.44  \\
    {+ Scene-CoT} &  & \underline{57.49}  & 50.73  & \underline{52.42} & \underline{55.00}  & 30.77  & 48.91  & \underline{45.65}  & \underline{48.71}  \\
    {+ TCogMap} & & \textbf{62.79} & \textbf{65.79} & \textbf{50.81} & \textbf{60.00} & \textbf{46.26} & \textbf{69.57} & \textbf{57.95} & \textbf{62.18}  \\
    \rowcolor{GroupBlue} \textit{Qwen2.5-VL-32B}~\cite{qwenvl} & Gen & 60.56 & 53.25 & 47.58 & 56.25 & \textbf{44.96} & \underline{60.87} & \textbf{62.91} & 55.20 \\
    \rowcolor{GroupBlue} \textit{Qwen2.5-VL-32B+EP} & Gen & \underline{64.13} & \underline{59.24} & \underline{53.23} & \underline{60.00} & \underline{44.45} & 52.17 & \underline{60.10} & \underline{56.19} \\
    {+ Scene-CoT} &  & 60.45  & 54.09  & 46.77 & 55.00  & 31.56  & 44.57  & 48.99 & 48.78  \\
    {+ TCogMap} &  & \textbf{71.85} & \textbf{68.59} & \textbf{58.87} & \textbf{65.00} & 41.42 & \textbf{78.26} & \textbf{62.91} & \textbf{63.84} \\ \midrule
    \rowcolor{GroupBlue} \textit{InternVL3-8B}~\cite{internvl3} & Gen & 57.26 & \underline{59.85} & 53.23 & 50.00 & 35.42 & \underline{52.17} & 45.64 & 50.51 \\
    \rowcolor{GroupBlue} \textit{InternVL3-8B+EP} & Gen & 53.57 & 57.28 & \underline{54.84} & 47.50 & \underline{37.58} & 46.74 & \underline{62.96} & 51.50 \\
    {+ Scene-CoT} &  & \textbf{62.52}   & 54.31   & 51.61 & \underline{56.25} & 37.31  & 51.09  & 50.34  & \underline{51.92}  \\
    {+ TCogMap} &  & \underline{61.37} & \textbf{66.69} & \textbf{57.26} & \textbf{66.25} & \textbf{47.08} & \textbf{67.39} & \textbf{64.10} & \textbf{61.45} \\
    \rowcolor{GroupBlue}\textit{InternVL3-14B}~\cite{internvl3} & Gen & 56.87 & 53.70 & \underline{52.42} & \underline{56.25} & 43.53 & \underline{59.78} & 43.90 & 52.35 \\
    \rowcolor{GroupBlue} \textit{InternVL3-14B+EP} & Gen & 60.41 & \underline{59.80} & \underline{52.42} & 55.00 & \underline{46.78} & 52.17 & \textbf{62.83} & \underline{55.63} \\
    {+ Scene-CoT} &  & \underline{62.06} & 53.42  & 51.61 & \underline{56.25}  & 36.61  & 51.09  & 51.64 & 51.81  \\
    {+ TCogMap} &  & \textbf{66.09} & \textbf{65.29} & \textbf{58.06} & \textbf{71.25} & \textbf{50.24} & \textbf{75.00} & \underline{59.00} & \textbf{63.56} \\
    \rowcolor{GroupBlue} \textit{InternVL3-38B}~\cite{internvl3} & Gen & 50.69 & 58.34 & 45.97 & 55.00 & \textbf{45.41} & \underline{71.74} & \textbf{68.27} & 56.49 \\
    \rowcolor{GroupBlue} \textit{InternVL3-38B+EP} & Gen & 55.57 & \underline{61.81} & 52.42 & 56.25 & 38.33 & 68.48 & 66.00 & \underline{56.98} \\
    {+ Scene-CoT} &  & \underline{63.67}  & 55.77  & \underline{53.23}  & \underline{57.50}  & 38.27  & 55.43  & 53.77  & 53.95  \\
    {+ TCogMap} &  & \textbf{67.01} & \textbf{68.25} & \textbf{60.48} & \textbf{72.50} & \underline{44.51} & \textbf{80.43} & \underline{66.43} & \textbf{65.66} \\ \midrule \midrule
    \rowcolor{GroupBeige} \textit{RoboTron}~\cite{RoboTron} & Drv & 43.63 & 58.17 & 29.84 & 41.25 & \underline{31.00} & \underline{65.22} & 28.92 & 42.58 \\
    \rowcolor{GroupBeige} \textit{RoboTron+EP} & Drv & 43.01 & \underline{59.69} & 33.06 & 43.75 & 28.01 & 61.96 & \underline{36.10} & 43.65 \\
    {+ Scene-CoT} &  & \textbf{58.68}  & 52.80  & \textbf{52.42} & \textbf{57.50}   & 30.40   & 54.35  & 34.79   & \underline{48.71}  \\
    {+ TCogMap} &  & \underline{48.20} & \textbf{62.77} & \underline{37.90} & \underline{52.50} & \textbf{40.15} & \textbf{79.35} & \textbf{40.58} & \textbf{51.64} \\ \midrule
    \rowcolor{GroupBeige} \textit{Cosmos-Reason}~\cite{cosmos} & Drv & 50.46 & 49.38 & 39.52 & 52.50 & 16.39 & \underline{64.13} & 16.06 & 41.21 \\
    \rowcolor{GroupBeige} \textit{Cosmos-Reason+EP} & Drv & 47.08 & \underline{53.86} & 37.90 & 52.50 & \textbf{37.42} & 45.65 & \textbf{46.70} & 45.87 \\
    {+ Scene-CoT} &   & \underline{54.38} & 47.87 & \textbf{55.65} & \underline{60.00} & 28.79 & 55.43 & \underline{43.72} & \underline{49.41} \\
    {+ TCogMap} & & \textbf{58.26} & \textbf{64.61} & \underline{50.00} & \textbf{67.50} & \underline{36.39} & \textbf{67.39} & 23.97 & \textbf{52.59} \\
        \bottomrule
    \end{tabular}
\label{table_main_results}
\end{table*}

\subsection{Experimental Setup}
For most TAD benchmarking results, four variants of each model are presented: baseline, baseline + textual ego pose, Scene-CoT (baseline + CoT outputs), and TCogMap (baseline + Temporal Cognitive Map). For the baseline, the VLM receives only the scene frames (approximately 40) and the question. The two segment tasks, \textit{Exact Answer Action Recognition} and \textit{Multiple Choice Action Recognition}, are exceptions and instead utilize approximately ten frames from the relevant segment as input to the VLMs. The baseline + textual ego pose variant is included for fair comparison with TCogMap.

Most questions in TAD are multiple choice and are evaluated using accuracy. Performance for questions that output a frame list is measured via temporal mean intersection over union (mIoU), a standard metric for temporal localization~\cite{LITA}. For questions expecting output text, a binary exact-match metric is used.

For Scene-CoT, Qwen2.5-14B-Instruct-1M~\cite{qwen2.5} was used as the LLM, unless specified otherwise. For TCogMap, the thresholds were empirically set to $V_{\text{stat}}=0.2$\,m/s, $V_{\text{stopping}}=1.0$\,m/s, $\Psi_{\text{turn}}=10.0^\circ$, $V_{y, \text{lc}}=0.4$\,m/s, $V_{x, \text{lc}}=1.0$\,m/s.

\subsection{Chance and Human Performance}
Human and chance performance were also computed on TAD. Chance performance was computed based on random selection for multiple-choice questions; a random subset of frames was used for temporal localization tasks. Human-level performance was computed using a randomly-sampled $10\%$ subset of each question type. The first two rows in \cref{table_main_results} show human-level and chance performance. Human performance is well above chance across the tasks. Performance on \textit{Temporal Action Localization} is somewhat lower for humans, likely due to the similarity of some motion categories (e.g., stopped vs.\ stopping) and minor misalignments between the ground truth (in segments) and human responses (in frames).

\subsection{Benchmarking Generalist VLMs}
Experiments were performed using both closed- and open-source generalist VLMs not specifically trained for AD. For closed-source models, GPT-5-mini~\cite{gpt5} and Gemini-2.5-Flash~\cite{gemini} were tested. For open-source models, Qwen2.5-VL~\cite{qwenvl} and InternVL3~\cite{internvl3} were selected, as these families have consistently provided top-tier performance and have the added benefit of released checkpoints across a range of sizes. Further parameter details are provided in the supplementary materials.

\textbf{Open-Source Models.}
As shown in \cref{table_main_results}, for the baseline models, the average performance generally increases with the model size: Qwen2.5-VL improves by more than 10\% from 7B to 32B. Scene-CoT, the proposed CoT-based method, provides modest gains over the baselines for smaller generalist models (e.g., Qwen2.5-VL-7B with Scene-CoT yields a 4.25\% improvement). For larger models, Scene-CoT does not offer additional gains; this result is hypothesized to arise from larger VLMs' stronger inherent capabilities, which let them produce internal representations better suited for temporal QA than an explicit CoT-based description.

TCogMap consistently outperforms the \textit{baseline + ego pose} variants in average performance by as much as 17.74\% (Qwen2.5-VL-7B), and large models still show substantial gains (e.g., 8.68\% with InternVL3-38B). Even though baseline + ego pose and TCogMap receive identical raw information, the VLM is unable to digest the raw pose data, illustrating the value of the structured temporal cognitive map.

\textbf{Closed-Source Models.} The average performance of GPT-5-mini and Gemini-2.5-Flash is comparable to the medium-sized InternVL3-14B. Interestingly, GPT-5-mini shows strong results on two of the ``easier'' tasks, \textit{Video Temporal Ordering} and \textit{Relative Temporal Localization}. Gemini-2.5-Flash performs worse on these two tasks, but achieves higher accuracies for \textit{Temporal Action Localization} and \textit{Temporal Object Localization}.

\subsection{Benchmarking on AD-Specific Models}
Two models trained on AD data, RoboTron (a.k.a.\ DriveMM)~\cite{RoboTron} (8B) and Cosmos-Reason~\cite{cosmos} (7B), were selected as the specialist models, due to their recent SoTA results and demonstrated ability to perform a range of AD tasks. As shown in \cref{table_main_results}, the baselines for these models are on par with Qwen2.5-VL-7B. However, Cosmos-Reason is significantly outperformed by RoboTron on the \textit{Temporal Action Localization} and \textit{Temporal Object Localization} tasks, likely due to its input design: Cosmos-Reason receives entire videos rather than frame sequences, and thus lacks the explicit frame-index awareness required for these temporally-grounded questions.

These results indicate that in-domain training data is not a panacea for improving temporal understanding. Beyond the influence of training data, mechanisms that explicitly capture temporal structure and causal dependencies across frames are required to enhance understanding. Incorporating the proposed Scene-CoT and TCogMap as training-free extensions on top of the AD-specific models results in average accuracy gains of up to 11.38\% on TAD, a directly actionable improvement for any agentic driving system that uses these VLMs as components.

\subsection{Ablation Studies}
Ablations use InternVL3-8B as the base model unless noted; further ablations (LLM choice, inference time) are in the supplementary materials.

\textbf{Ego vs.\ Non-ego Performance Gains.} As mentioned in \cref{sec:tad-benchmark}, the questions related to the ego and non-ego vehicles are relatively balanced; however, it is crucial to investigate whether the performance varies between these two question categories. \cref{table_ego_vs_non_ego} shows the average accuracies for ego and non-ego questions. 
The baseline VLM achieves higher accuracy for ego-related questions. Ego-related cues are implicitly encoded in every frame through camera motion, whereas non-ego agents appear intermittently at lower resolution, giving the model a longer, more stable signal for ego questions, yielding better accuracies.

Furthermore, while Scene-CoT drops accuracy slightly for ego questions (0.52\%) compared to the baseline (see \cref{table_ego_vs_non_ego}), it improved performance on non-ego questions by more than 5\%. Finally, results for TCogMap show boosts of 13.02\% and 4.57\% on ego and non-ego questions, respectively. These results are encouraging as they show that the additional context of the ego's motion offers the VLM with hints about other objects' actions and generally improves the model's temporal understanding of the scene. 
Further analysis with other models and detailed task accuracies for ego vs. non-ego are given in the supplementary materials.

\begin{table}[t!]
    \centering
    \caption{Average accuracies on ego vs.\ non-ego questions.}
    \label{table_ego_vs_non_ego}
    \begin{tabular}{l|cc}
        \toprule
        \textbf{Method} & \textbf{Ego Qs} & \textbf{Non-Ego Qs} \\
        \midrule
        \rowcolor{GroupBlue} \textit{InternVL3-8B} & 55.15 & 43.24\\
        \quad + Scene-CoT & 54.63 & 48.62 \\
        \quad + TCogMap & 68.17 & 47.81\\
        \bottomrule
    \end{tabular}
\end{table}

\textbf{Blind VLM Test.} This ablation examines two questions about the TAD benchmark: 1) How much a model can infer from the question alone. 2) Whether temporal cognitive maps are sufficient for temporal reasoning in isolation.

\begin{figure}[t]
    \centering
    \includegraphics[width=0.8\columnwidth]{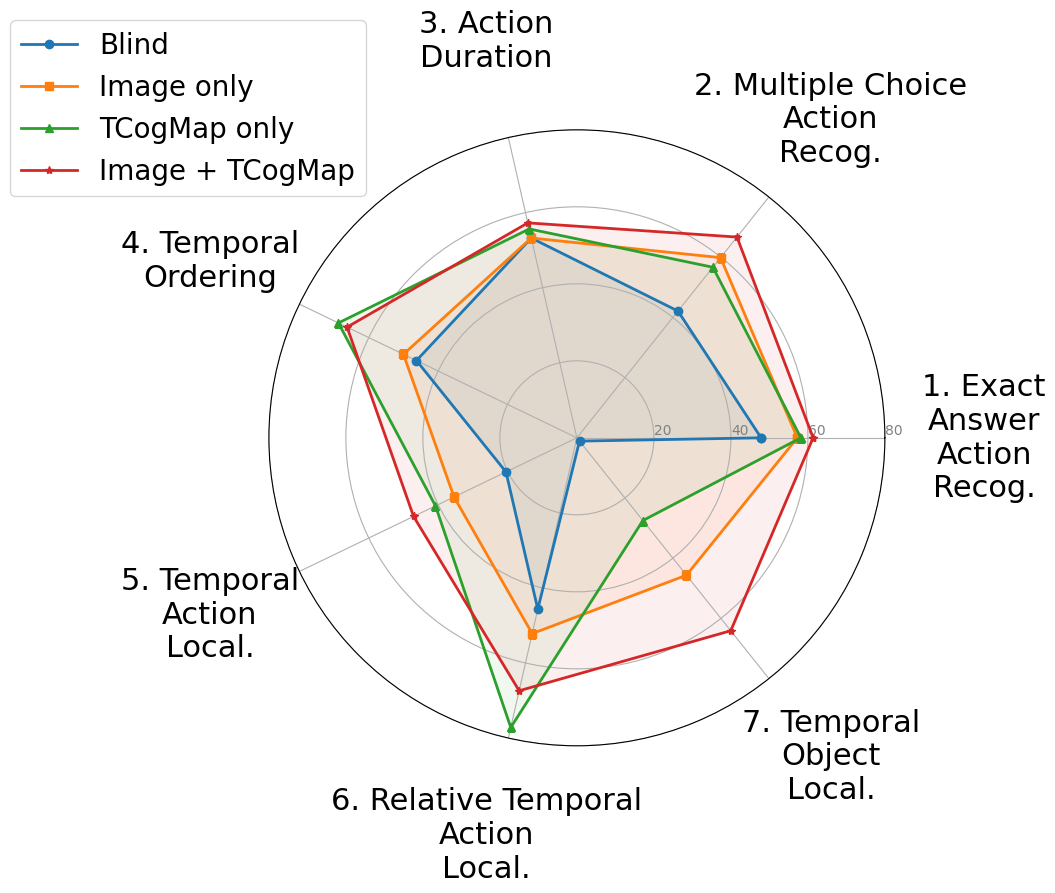}
    \caption{Blind test results for InternVL3-8B across the four configurations (Blind, Image Only, TCogMap Only, Image + TCogMap).}
    \label{fig:blind_radar}
\end{figure}

As shown in \cref{fig:blind_radar}, four configurations were evaluated: \textbf{1)~Blind}: Only the question is provided to the VLM. \textbf{2)~Image Only:} Scene frames and question are given as input. \textbf{3)~TCogMap Only:} Only the temporal cognitive map and question are provided, excluding frames. \textbf{4)~Image + TCogMap:} All inputs (frames, TCogMap, and question) are provided together. 
Task accuracies in the \textbf{Blind} setting are the lowest among all configurations, often approaching chance-level. 
Comparing \textbf{Blind} with \textbf{Image Only} confirms that the benchmark questions cannot be answered without visual input. Interestingly, using the \textbf{TCogMap Only} setting yields higher accuracies in five tasks compared to \textbf{Image Only}.
This result demonstrates that the proposed {TCogMap} effectively captures temporal dynamics even without direct visual input. Finally, combining both frames and {TCogMap} achieves best performance across most tasks, indicating that while {TCogMap} substantially enhances temporal reasoning, visual content remains essential for complete temporal understanding. 


Scene-CoT and TCogMap were minimally adapted for STSBench to account for the shorter segment lengths, and the results are shown in \cref{table_sts_bench}. The result trends are similar to those seen in \cref{table_main_results}. In particular, Scene-CoT provides a modest average improvement over the base VLM, indicating the merits of extracting an intermediary CoT reasoning trace and allowing an LLM to perform final question-answering. TCogMap also yields an impressive boost of more than 10\% over the base model. As expected, TCogMap's most significant gains are for ego vehicle questions ({Ego} and {Ego-Agent} tasks), but it still provides modest improvements for non-ego tasks. Similar to the TAD findings, this result reinforces that even if a question is not directly related to the ego vehicle, the context information provided by the ego's trajectory can still be instructive. 
\begin{table}[t]
    \centering
    \caption{Task and average accuracies on STSBench for the InternVL3-8B baseline and the proposed methods.}
    \label{table_sts_bench}
    \begin{tabular}{l|cccc|c}
        \toprule
        \textbf{Method} & \textbf{Ego} & \textbf{Ego-} & \textbf{Agent} & \textbf{Agent-} & \textbf{Avg.} \\
        & & \textbf{Agent} & & \textbf{Agent} & \\
        \midrule
        \rowcolor{GroupBlue} \textit{InternVL3-8B} & 34.36 & \underline{51.41} & \underline{46.69} & 39.52 & 43.00  \\
        \quad + Scene-CoT & \underline{40.05} & 50.83 & 37.51 & \textbf{51.38} & \underline{44.94} \\
        \quad + TCogMap & \textbf{56.49} & \textbf{61.45} & \textbf{52.94} & 42.53 & \textbf{53.35} \\
        \bottomrule
    \end{tabular}
\end{table}
Overall, the results for Scene-CoT and TCogMap show that they can generalize and provide accuracy increases on other datasets.

\textbf{Generalization to STSBench.} Scene-CoT and TCogMap were also evaluated on STSBench~\cite{stsbench}, the most closely-related AD benchmark, which contains only short (3-second) event-recognition segments. With minor adaptations, Scene-CoT yields a modest average gain over the InternVL3-8B baseline, while TCogMap yields a 10.35\% boost (43.00\%~$\rightarrow$~53.35\% average), with the largest gains on ego-centric tasks --- evidence that the agentic-style augmentation transfers across temporal AD benchmarks. Per-task results are in the supplementary materials.

\section{Conclusion and Future Work}
This work introduced TAD, a benchmark for temporal understanding in autonomous driving, a domain that exemplifies the challenges of deploying agentic AI in the real world. Evaluation of generalist and AD-specific VLMs revealed a substantial gap between current SoTA models and human performance, directly limiting the reliability of any agent that uses such VLMs as components. To address this, two training-free solutions were proposed: \textbf{Scene-CoT}, a CoT-based reasoning approach, and \textbf{TCogMap}, which leverages a deterministic trajectory-analysis tool to feed the VLM an ego-centric temporal cognitive map. Both approaches follow a lightweight, tool-augmented agentic pattern that is increasingly central to making autonomous agents reliable in open-ended environments. Future work includes extending temporal cognitive maps to non-ego objects, integration into closed-loop planning agents, and creation of a fine-tuning dataset for VLM-based agentic AD systems.

\newpage
\section*{Impact Statement}
This paper presents work whose goal is to advance the field of Machine Learning, 
specifically in the context of Vision-Language Models (VLMs) for autonomous driving (AD). 
By introducing the Temporal Understanding in Autonomous Driving (TAD) benchmark and 
proposing training-free methods to improve temporal reasoning, this work aims to 
contribute to the development of safer and more reliable autonomous driving systems. The societal implications of this work are predominantly positive: improved temporal 
understanding in VLM-based driving agents can directly contribute to safer autonomous 
vehicles, with downstream benefits for road safety and transportation efficiency. 
However, we acknowledge that advances in autonomous driving perception and reasoning 
carry broader responsibilities. Overestimating the capabilities of such systems based 
on benchmark performance---without accounting for real-world complexity, edge cases, 
and distribution shift---poses safety risks if results are prematurely translated into 
deployment. We therefore caution that TAD, while a rigorous testbed, should be 
regarded as one component of a broader validation framework rather than a definitive 
measure of deployment readiness. Regarding data, TAD is built upon publicly available datasets (e.g., NuScenes), and 
our human-annotated action labels were collected under standard annotation protocols. 
No personally identifiable information beyond what is already present in the source 
datasets is introduced. We encourage future users of TAD to adhere to the licensing 
terms of the underlying data sources. In summary, we believe the benefits of this work---providing a principled benchmark 
and effective enhancements for temporal reasoning in safety-critical agentic 
systems---outweigh its risks, provided that the limitations described herein and in 
the main paper are carefully observed by practitioners.


\bibliographystyle{icml2026}
\bibliography{main}

\clearpage
\appendix
\section{TAD Benchmark: Details and Additional Statistics}
In the main body of the paper, fundamental statistics regarding the benchmark, such as the number of questions per task, are provided.  Further analyses on the statistics of the TAD benchmark are provided in this section. 

\subsection{Vehicle Action Statistics in TAD}
\begin{figure}[h]
    \centering
    \includegraphics[width=0.9\columnwidth]{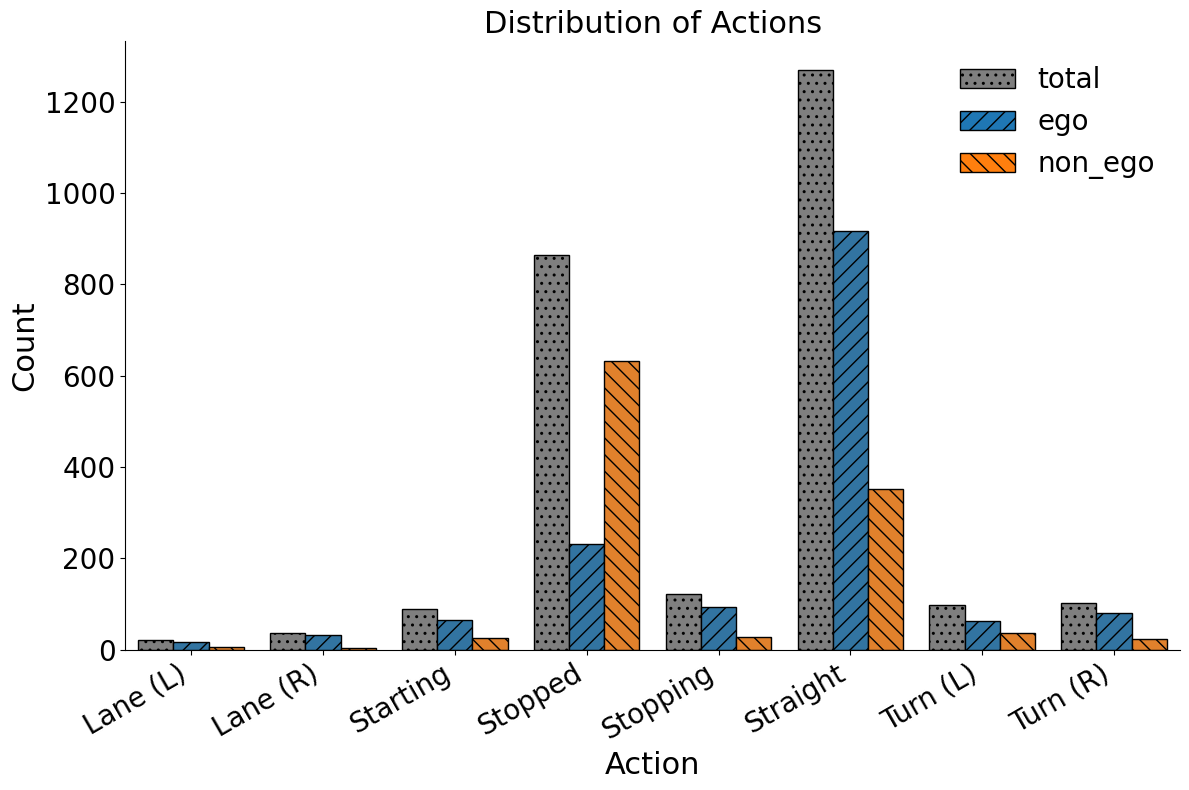}
    \caption{Distribution of the actions in the \textit{Exact Answer Action Recognition} question type. Shorthand notations: 
    Lane (L): change lane to left, Lane (R): change lane to right, Turn (L): Turn left, Turn (R): Turn right, and Straight: Straight, constant speed }
\label{fig:action_barplot}
\end{figure}

To determine the distribution of vehicle actions in the benchmark, an analysis was performed on the \textit{Exact Answer Action Recognition} question type, the results of which are shown in \cref{fig:action_barplot}.
The data is presented in terms of ego, non-ego, as well as the combined total. 
The two most dominant action categories are ``stopped'' and ``straight, constant speed''. Notably, the ego vehicle is most commonly found to be traveling ``straight, constant speed''; whereas, non-ego vehicles are most often stopped.
This result is quite intuitive, as the NuScenes data generally shows the ego vehicle driving throughout cities with a significant number of stationary vehicles parked in lots or along streets. 
Thus, these statistics reflect the underlying {NuScenes data distribution.}
{Note that in the current work,} action labels were also assigned to the video segments for the first time. A total of 4,481 vehicle actions were labeled during the human annotation procedure.

\subsection{Vehicle Action Definition Clarification}
To avoid confusion, an additional note should be made regarding the definition of the actions used in TAD. In particular, the actions are defined \textbf{with respect to the vehicle that is performing the action.} As an example, imagine a scenario where a truck is traveling in the opposite direction to the ego vehicle and subsequently, the truck turns right onto a side street. In TAD, the truck's action would be labeled as  \textbf{Turn Right}, since the driver of the truck has literally turned the wheel of his/her vehicle to make a right hand turn. The labels in TAD are \textbf{not} based on how the ego vehicle observes vehicles moving from its own perspective (i.e., in this example, within the ego vehicle's front camera field of view, the truck would be seen as traveling further to the left during its turn). Defining vehicle actions in this manner presents an additional challenge for VLMs, but models that can differentiate between such actions exhibit a greater understanding of the environment, road layout, and 3D scene. 

\subsection{Word Cloud of Questions in the TAD Benchmark}
\begin{figure}[t]
    \centering
    \includegraphics[width=0.9\columnwidth]{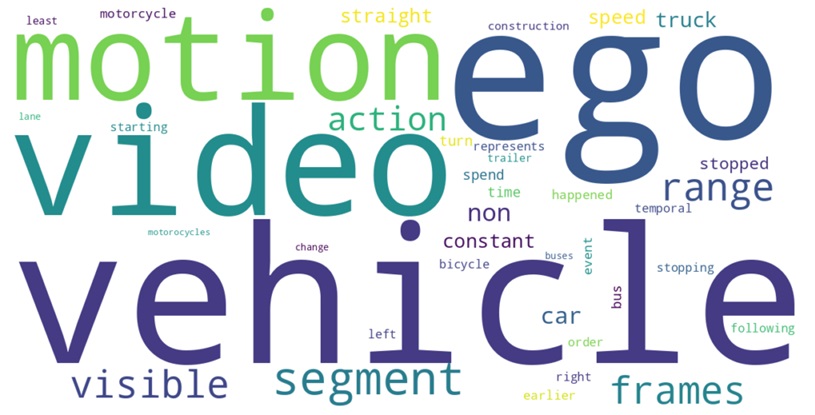}
    \caption{Word cloud for the questions in the TAD benchmark (excluding the instructions)}
\label{fig:action_wordcloud}
\end{figure}

The word cloud in \cref{fig:action_wordcloud} summarizes the distribution of question terms in the benchmark after removing instructional tokens related to the answer format. The most prominent words are vehicle, motion, ego, video, segment, and action, which highlight the benchmark’s emphasis on reasoning about temporal dynamics. The dominance of these words reflects the centrality of motion cues, ego-vehicle context, and temporal segmentation in the tasks, underscoring the benchmark’s focus on temporal understanding in autonomous driving.

\section{Parameters for the Baselines}
In this section, the specific parameters and settings related to the baselines in the main body are introduced.
\begin{itemize}
    \item \textbf{GPT-5-mini:} The model is called using its default API configuration, with reasoning enabled by default and set to ``medium''. \texttt{max\_new\_token} is set to 2048. 
    \item \textbf{Gemini-2.5-Flash:} The model is called using its default API configuration, with thinking enabled by default and the thinking budget set to 1024 tokens. Gemini-2.5-Flash supports video inputs. Hence, the sampled frames in the NuScenes dataset are stored as videos with FPS=2. Then, the stored videos are used for inference. \texttt{max\_new\_token} is set to 2048. 
    \item \textbf{InternVL3 family:} The LMDeploy inference framework~\cite{lmdeploy} is used to get the inference results for the InternVL3 family of models. To keep the number of visual tokens produced by long sequences of frames within the model's maximum context length, \texttt{max\_dynamic\_patch=1} is used. \texttt{max\_new\_token} = 1024 is used for this model.  
    \item \textbf{Qwen-2.5-VL family:} The LMDeploy inference framework~\cite{lmdeploy} is used to get the inference results for Qwen-2.5-VL family of models. To keep the number of visual tokens obtained from long sequences of frames within the model's maximum context length, \texttt{min\_pixels}=$256 \times 28 \times 28$ and \texttt{max\_pixels}=$512 \times 28 \times 28$ are used. \texttt{max\_new\_token} = 1024 is used for this model.  
    \item \textbf{Robotron:} The parameters and prompts provided in their official examples are used~\footnote{\url{https://github.com/zhijian11/RoboTron-Drive/blob/main/scripts/inference_demo/demo_video.py}}.
    \item \textbf{Cosmo-reason:} The parameters provided in their official  code~\footnote{\url{https://github.com/nvidia-cosmos/cosmos-reason1/blob/main/scripts/inference_sample.py}} are used. Since the model receives video as input, the sampled frames in the NuScenes dataset are converted to videos with FPS=2. The stored videos are then used for inference. 
\end{itemize}

\subsection{Task Details and Examples}
In the main body of the paper, Fig. 1 
provides an overview of the TAD benchmark, including a listing of the seven different question types, as well as the frames and labels for several video segments. In this section, additional details and examples from TAD are given.

\cref{tab:supp_task_desc} provides a more detailed description for each of the seven question types, along with an explicit example (including both the question and answer) from the benchmark. In the description, some of the motivations that inspired the inclusion of question types are also listed. 

\newcolumntype{L}[1]{>{\raggedright\arraybackslash}p{#1}}

\begin{table*}[tb!]
\caption{TAD Question Types and Examples}
\label{TaskTable}
\tiny
\begin{tabular}{  L{2cm} | p{7.1cm} | p{7.1cm} }
    \toprule
    \textbf{Task}      
    & \textbf{Description}   
    & \textbf{Example} \\\midrule
    Exact Answer Action Recognition
    & Evaluates if a model can perform vehicle action recognition, on a short, temporally cropped video segment. This segment-level question type isolates the VLM's ability to perform fine-grained action recognition without concerning longer term temporal understanding in a full-length video. The model is expected to produce the exact textual answer from a list of action options.        
    & \textbf{Question:} What best describes the motion of the bus visible in this video segment? Respond with exactly one full phrase from the following list: `Starting', `Stopping', `Turn left', `Turn right', `Change lane to the left', `Change lane to the right', `Straight, constant speed', `Stopped'

    \textbf{Answer:} \textit{Turn left}    

    \\\hline
    Multiple Choice Action Recognition   
    & Evaluates if a model can understand the action of a vehicle in a short, temporally cropped video segment. This segment-level question type isolates the VLM's ability to perform fine-grained action recognition without concerning longer term temporal understanding in a full-length video. 
    & \textbf{Question:} What is the motion of the ego vehicle in this video segment? Respond with exactly one letter corresponding to the correct option. A. Turn left. B. Straight, constant speed. C. Stopped. D. Starting.

    \textbf{Answer:} \textit{D}

    \\\hline
    
    Action Duration       
    & Assesses if a model can understand the durations of various events that transpire in a video. The question asks which event lasts the longest/shortest in a video and is presented with several options in a multiple choice setting. The actions for this question type are all related to the ego vehicle.  
    & \textbf{Question:} Which action did the ego vehicle spend the most time doing? Respond with exactly one letter corresponding to the correct option. A. Straight, constant speed. B. Starting. C. Stopping. D. Stopped.
    
    \textbf{Answer:} \textit{A}
    \\\hline
    Temporal Ordering &
    Determines if a model can properly identify the temporal sequence of actions performed by a specific vehicle in the scene. Various action sequences are presented as multiple choice options.        
    & \textbf{Question:} Which of the following represents the correct temporal order of motions for the ego vehicle? Respond with exactly one letter corresponding to the correct option. A. Starting; Straight, constant speed; Stopping. B. Stopping; Starting; Straight, constant speed; C. Starting, Stopping; Straight, constant speed; D. Stopping; Straight, constant speed; Starting.    
    
    \textbf{Answer:} \textit{C}    
    \\\hline   

    Temporal Action Localization &
    Evaluates if a model can identify the frames where a specific action is being performed by a vehicle. The expected output is a list of the frames that correspond to the specified actions.
    
    & \textbf{Question:} In which frames is the ego vehicle doing action `stopped'? Respond with an explicit list of all frame numbers, enumerating each value individually, without summarizing as a range.
    
    \textbf{Answer:} \textit{[0, 1, 2, 3, 4, 5, 6, 7, 8, 9]}
    \\\hline

    Relative Temporal Action Localization &
    Assesses if a model can identify which of two vehicle actions happened earlier/later in the video. Whereas temporal localization exactly localizes a single action, this task requires localization of two actions and determining their relative temporal ordering. The action answers are presented as multiple choice options.       
    & \textbf{Question:} Which ego vehicle event happened earlier in the video? Respond with exactly one letter corresponding to the correct option. A. Stopping. B. Starting.
    
    \textbf{Answer:} \textit{B}    
    \\\hline     

    Temporal Object Localization &
    Evaluates if a model can identify the frame ranges where a specific object is present. This task disentangles the challenges of action recognition and temporal localization (i.e., to determine if the VLM can identify the frames where a \textbf{non-action} event occurs in a video). The expected output is a list of the frames that correspond to the question criteria.
    & \textbf{Question:} In which frames are there any trucks visible to the ego vehicle? Respond with an explicit list of all frame numbers, enumerating each value individually, without summarizing as a range.
    
    \textbf{Answer:} \textit{[1, 2, 3, 4, 5, 6, 7, 8, 9, 10, 11, 12, 13, 14, 15, 16, 17]}

    \\\bottomrule
\end{tabular}
\label{tab:supp_task_desc}
\end{table*}

\begin{figure*}[tb!]
     \centering     \includegraphics[width=1.0\textwidth]{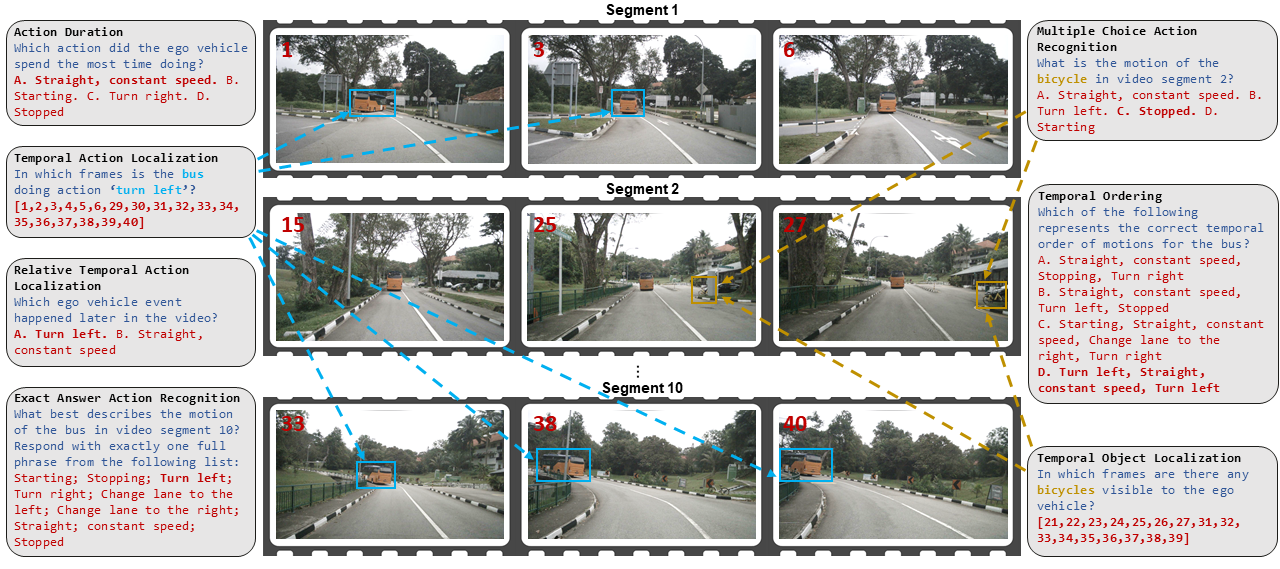}
     \caption{Pictorial visualization example of the question types and format in the TAD benchmark.}
    \label{fig:QOverview}
\end{figure*}

An additional pictorial example showing the format of questions in TAD is provided in \cref{fig:QOverview}. This figure is meant for visualization purposes only to understand the format of the benchmark questions. The figure highlights the different question types and the breakdown of the video into segments. The seven question types are shown in grey bubbles. Dark blue text corresponds to the questions, while red text shows the 
{available answer options (note, not all of the options are \textbf{correct})}
and the correct answers are \textbf{bolded}. Three video segments are displayed (Segment 1, Segment 2, and Segment 10), represented by three frames from the respective segment. Cyan text and arrows are used to highlight questions related to the actions of the bus; whereas orange text and arrows draw attention to the bicycle.  The frame numbers displayed on the image in red (top-left corner) are also included only for visualization. In practice, the red numbers are not superimposed on the images that are used as input to the VLMs. 

  \begin{figure*}[t!]
 \centering
  \includegraphics[width=0.95\textwidth]{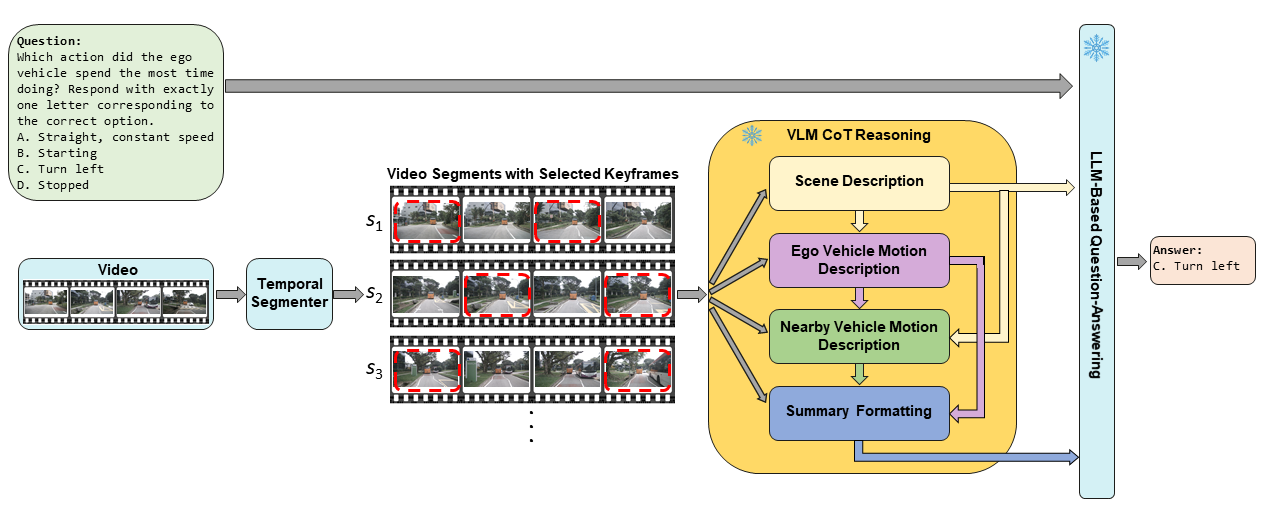}
 \caption{Overview of the proposed Scene-CoT method.}
 \label{fig:Cap4Time}
 \end{figure*}

\section{Scene-CoT: Additional Details}


In the main body of the paper, the critical details of 
the proposed Scene-CoT method were provided. 
Indeed, the inspiration of the Scene-CoT method was to design a general baseline that divides the QA problem into a VLM-based video perception module and an LLM-based reasoner. There is related ongoing research that shows the benefits of such perception and reasoning decoupling for solving complex math and science problems \cite{gou2025perceptual,guo2025decoupled} as well as for RL-based finetuning \cite{rezaei2026cppo}.
Here, more details of the Scene-CoT approach, including the overall system diagram (shown in \cref{fig:Cap4Time}), are provided. Of note, the VLM CoT reasoning block shows the information flow during the series of CoT VLM calls. As discussed in the main paper, general scene descriptions are first generated using a uniformly sampled sub-set of frames in each segment. Using these scene descriptions as additional context, the VLM is prompted to describe the ego vehicle motion and subsequently the nearby vehicles' motion. The VLM also receives the additional context of the ego vehicle motion descriptions, as well as the scene descriptions, when it is prompted to describe the motion of the nearby vehicles. The final step in the CoT generates a JSON-formatted output summary using the ego and nearby motion descriptions as contextual information. Finally, to perform question-answering, the LLM is provided with three inputs: 1) The original question. 2) The general Scene Description. 3) The JSON-style motion summary.

\begin{figure*}[tb!]
 \centering
  \includegraphics[height=0.85\textheight,keepaspectratio]{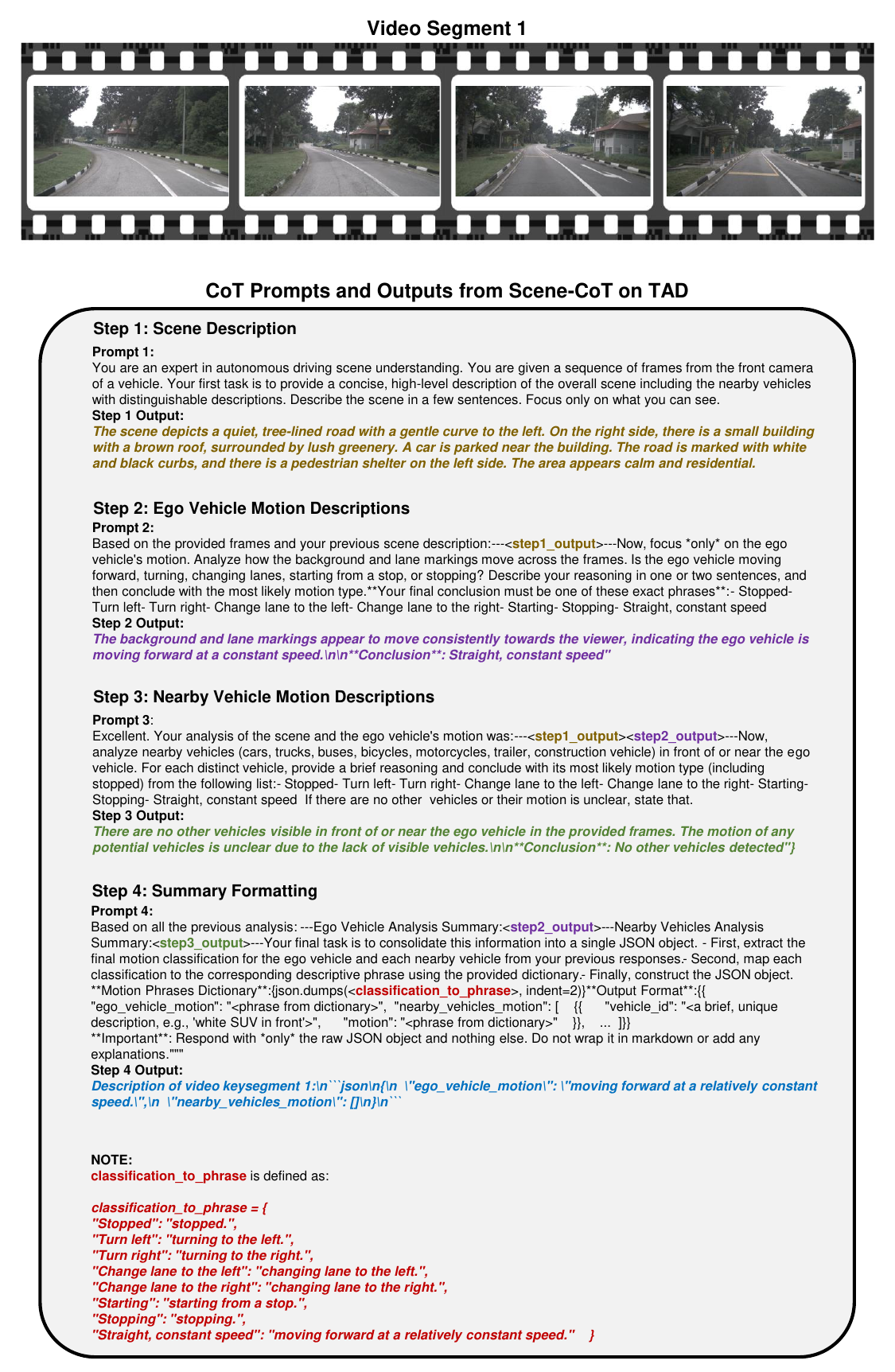}
\vspace{-4pt}
 \caption{Exact prompts and sample output with Scene-CoT for a question in TAD.}
 \label{fig:cot_prompt_ans} 
 \end{figure*}

A concrete example from TAD showing the series of CoT prompts, along with the VLM's outputs is shown in \cref{fig:cot_prompt_ans}.

\section{Additional Ablation Studies}
In the main body of the paper, ablations were presented that showed accuracy breakdowns for ego and non-ego questions, a Blind VLM test, as well as performance generalization on STSBench \cite{stsbench}. In this section, results for five additional ablation experiments are presented. 
The ablation experiments are as follows: 1) \textbf{Ego vs. Non-Ego Results.} Additional details and results further analyzing model accuracies on ego and non-ego questions. 2) \textbf{Captioning Style.} Ablations on the prompting style used in Scene-CoT, for creating the segment descriptions. 3) \textbf{QA LLM Selection.} An ablation experiment considering the effect of the specific LLM used in Scene-CoT. 4) \textbf{Object detections as context}. An ablation that considers using the output of an object detector to improve captioning quality.  5) \textbf{Inference Time.} An analysis of inference time for the methods. 

\subsection{Ego. vs Non-Ego Performance Gains: Additional Results and Analysis}
As mentioned in the main body of the paper, the questions related to the ego and non-ego vehicles are relatively balanced; however, it is still important to consider whether the performance varies between these two question categories. 

In the main body of the paper, Tab. 5 shows a summarized version of the results for the ego vs. non-ego vehicle questions by presenting the \textbf{average} accuracies across all question types for the InternVL3-8B model. Here, these results are significantly expanded upon along two dimensions: 1) \textbf{Additional models.} The ego vs. non-ego breakdown is provided for six additional models (InternVL3-14B, InternVL3-38B, Qwen2.5-VL-7B, Qwen2.5-VL-32B, RoboTron, and Cosmos-Reason). This set of models forms the complete set of open-source generalist and specialist models that were considered in Tab 4 in the main body. 2) \textbf{Task-Wise Accuracies.} The ego vs. non-ego question accuracies are provided on a per-type basis, permitting a more fine-grained results analysis. The detailed ego vs. non-ego vehicle accuracies are presented in \cref{tbl:ego_q_main_results} and \cref{tbl:non_ego_q_main_results}. The format of these two tables matches that of the main results table in the body of the paper, but each table only evaluates a sub-set of the questions (i.e., ego questions in \cref{tbl:ego_q_main_results} and non-ego question in \cref{tbl:non_ego_q_main_results}). Note that “N/A” in a column indicates that the corresponding task does not include either ego-related or non-ego-related questions.

A comparison of the results in \cref{tbl:ego_q_main_results} and \cref{tbl:non_ego_q_main_results} shows that the impact of Scene-CoT varies notably across models. While some models, such as Qwen2.5-VL-7B, exhibit substantially larger gains on non-ego vehicle queries (+8.11\% vs. +3.38\% for ego), others display the opposite pattern. For instance, RoboTron benefits most on ego-centric questions (+10.81\%), suggesting that certain models are more capable of exploiting Scene-CoT’s structured reasoning for inferring ego-motion cues.

With TCogMap, \cref{tbl:ego_q_main_results} shows significant accuracy improvements on ego-related questions (typically around 15\%) and interestingly, more modest increases are still generally observed in \cref{tbl:non_ego_q_main_results} for non-ego questions. 
These results indicate that incorporating the ego vehicle’s motion offers useful contextual cues about the behavior of surrounding objects, leading to more reliable temporal reasoning by the VLM.



\renewcommand{\arraystretch}{0.90} 
\setlength{\tabcolsep}{5.0pt}
\begin{table*}[tb!]
    \caption{Results of \textbf{ego-related} questions for open- and closed-source generalist VLMs as well as AD-specific VLMs. 
    Bold and underline indicate best and second-best results when comparing the three system variants for each open-source VLM. \colorbox{GroupBlue}{Gen}: generalist models, \colorbox{GroupBeige}{Drv}: driving specialist models. \textbf{EA}: exact answer, \textbf{MC}: multiple choice.}
    \centering
    \begin{tabular}{w{l}{30mm}|w{c}{6.5mm}|w{c}{9.5mm}w{c}{9.5mm}w{c}{9.5mm}w{c}{9.5mm}w{c}{9.5mm}w{c}{9.5mm}w{c}{9.5mm}|w{c}{9.5mm}}
        \toprule
        \textbf{Method} 
        & \makecell{\textbf{Type}}  
        & \makecell{\textbf{EA }\\\textbf{Act.}\\\textbf{Recog.}} 
        & \makecell{\textbf{MC }\\\textbf{Act. }\\\textbf{Recog.}} 
        & \makecell{\textbf{Action}\\\textbf{Dur.}} 
        &\makecell{\textbf{Temp.}\\\textbf{Ord.}}
        & \makecell{\textbf{Temp. }\\\textbf{Act.}\\\textbf{Local.}} 
        & \makecell{\textbf{Rel. }\\\textbf{Temp.}\\\textbf{Local.}} 
        & \makecell{\textbf{Temp.}\\\textbf{Obj.}\\\textbf{Local.}} 
        & \textbf{Avg} \\ \midrule
    \rowcolor{GroupBlue} Qwen2.5-VL-7B & Gen &  63.53 & \underline{58.93} & 34.68 & \underline{56.45} & 35.00 & 46.74 & \text{N/A} & 49.22  \\
    {+ Scene-CoT} &  & \underline{65.13}  & 56.60   & \textbf{52.42}  & 54.84   & \underline{37.69}  & \underline{48.91}   & N/A   & \underline{52.60}  \\
    {+ TCogMap} & & \textbf{79.67} & \textbf{79.32} & \underline{50.81} & \textbf{62.90} & \textbf{48.87} & \textbf{69.57} & \text{N/A} & \textbf{65.19}  \\ 
    \rowcolor{GroupBlue} \textit{Qwen2.5-VL-32B} & Gen & 63.47 & 55.63 & \underline{47.58} & \underline{58.06} & \underline{46.09} & \underline{60.87} & \text{N/A} & \underline{55.81}  \\       
    {+ Scene-CoT} &  & \underline{67.93}  & \underline{60.58}   & 46.77  & 54.84   & 36.31  & 44.57  & N/A & 51.83  \\
    {+ TCogMap} &  & \textbf{80.53} & \textbf{78.35} & \textbf{58.87} & \textbf{74.19} & \textbf{46.23} & \textbf{78.26} & \text{N/A} & \textbf{69.83} \\  \midrule
    \rowcolor{GroupBlue} \textit{InternVL3-8B} & Gen & \underline{71.60} & \underline{62.33} & \underline{53.23} & 51.61 & 39.98 & \underline{52.17} & \text{N/A} & \underline{55.15} \\
    {+ Scene-CoT} & & 67.60  & 58.93  & 51.61  & \underline{56.45}  & \underline{42.08}  & 51.09   & N/A    & 54.63  \\
    {+ TCogMap} &  & \textbf{79.13} & \textbf{76.50} & \textbf{57.26} & \textbf{72.58} & \textbf{56.15} & \textbf{67.39} & \text{N/A} & \textbf{68.17} \\ 
    \rowcolor{GroupBlue}\textit{InternVL3-14B} & Gen & \underline{68.87} & \underline{60.87} & \underline{52.42} & \underline{59.68} & \underline{43.52} & \underline{59.78} & \text{N/A} & \underline{57.52}  \\
    {+ Scene-CoT} &  & 68.60 & 59.51   & 51.61  & 54.84  & 42.80   & 51.09  & N/A  & 54.74   \\
    {+ TCogMap} & & \textbf{80.53} & \textbf{77.67} & \textbf{58.06} & \textbf{80.65} & \textbf{55.83} & \textbf{75.00} & \text{N/A} & \textbf{71.29}  \\ 
    \rowcolor{GroupBlue} \textit{InternVL3-38B} & Gen & 55.67 & \underline{65.73} & 45.97 & 53.23 & \underline{46.15} & \underline{71.74} & \text{N/A} & 56.42  \\       
    {+ Scene-CoT} &  & \underline{70.00}  & 62.23   & \underline{53.23}   & \underline{56.45}   & 44.48   & 55.43   & N/A   & \underline{56.97}   \\
    {+ TCogMap} &  & \textbf{80.53} & \textbf{78.06} & \textbf{60.48} & \textbf{74.19} & \textbf{51.26} & \textbf{80.43} & \text{N/A} & \textbf{70.83}  \\ \midrule \midrule
      \rowcolor{GroupBeige} \textit{RoboTron} & Drv & 58.27 & \underline{71.26} & 4.84 & \underline{43.55} & 29.78 & \underline{65.22} & \text{N/A} & 45.49  \\
    {+ Scene-CoT} &  & \textbf{71.07}   & 64.17   & \textbf{52.42}  & \textbf{58.06}    & \underline{37.70}    & 54.35   & N/A   & \underline{56.30}   \\
    {+ TCogMap} &  & \underline{65.73} & \textbf{77.77} & \underline{37.90} & \textbf{58.06} & \textbf{46.46} & \textbf{79.35} & \text{N/A} & \textbf{60.88} \\ \midrule 
    \rowcolor{GroupBeige} \textit{
Cosmos-Reason} & Drv &  \underline{66.73} & \underline{61.55} & 39.52 & 54.84 & 16.66 & \underline{64.13} & \text{N/A} & 50.57 \\     
    {+ Scene-CoT} &  & 65.87 & 57.96 & \textbf{55.65} & \underline{59.68} & \underline{34.78} & 55.43 & N/A & \underline{54.90} \\
    {+ TCogMap} & & \textbf{80.13} & \textbf{79.71} & \underline{50.00} & \textbf{72.58} & \textbf{40.15} & \textbf{67.39} & \text{N/A} & \textbf{64.99}  \\
        \bottomrule
    \end{tabular}
\label{tbl:ego_q_main_results}
\end{table*}

\renewcommand{\arraystretch}{0.90} 
\setlength{\tabcolsep}{5.0pt}
\begin{table*}[tb!]
    \caption{Results of \textbf{non-ego} questions for open- and closed-source generalist VLMs as well as AD-specific VLMs. 
    Bold and underline indicate best and second-best results when comparing the 3 system variants for each open-source VLM. \colorbox{GroupBlue}{Gen}: generalist models, \colorbox{GroupBeige}{Drv}: driving specialist models. \textbf{EA}: exact answer, \textbf{MC}: multiple choice.}
    \centering
    \begin{tabular}{w{l}{30mm}|w{c}{6.5mm}|w{c}{9.5mm}w{c}{9.5mm}w{c}{9.5mm}w{c}{9.5mm}w{c}{9.5mm}w{c}{9.5mm}w{c}{9.5mm}|w{c}{9.5mm}}
        \toprule
        \textbf{Method} 
        & \makecell{\textbf{Model}\\\textbf{Type}}  
        & \makecell{\textbf{EA }\\\textbf{Act.}\\\textbf{Recog.}} 
        & \makecell{\textbf{MC }\\\textbf{Act. }\\\textbf{Recog.}} 
        & \makecell{\textbf{Action}\\\textbf{Dur.}} 
        &\makecell{\textbf{Temp.}\\\textbf{Ord.}}
        & \makecell{\textbf{Temp. }\\\textbf{Act.}\\\textbf{Local.}} 
        & \makecell{\textbf{Rel. }\\\textbf{Temp.}\\\textbf{Local.}} 
        & \makecell{\textbf{Temp.}\\\textbf{Obj.}\\\textbf{Local.}} 
        & \textbf{Avg} \\ \midrule
    \rowcolor{GroupBlue} \textit{Qwen2.5-VL-7B} & Gen &  26.36 & 40.61 & N/A           	& 27.78	& \underline{35.50}	 & N/A & 45.60 & 35.17  \\
     
    {+ Scene-CoT} &  & \textbf{47.10}  & \underline{42.72}  & N/A & \textbf{55.56}   & 25.35   & N/A   & \underline{45.65}  & \underline{43.28}  \\
    {+ TCogMap} & & \underline{39.86}	& \textbf{47.35}	&  N/A  & \underline{50.00} & \textbf{44.23} & N/A & \textbf{57.97}	& \textbf{47.88}  \\
    \rowcolor{GroupBlue} \textit{Qwen2.5-VL-32B} & Gen & \underline{56.61} &\underline{50.00}	& N/A &            	\underline{50.00}	& \textbf{44.08} & N/A  & \textbf{62.91}	& \textbf{52.72}  \\       
    {+ Scene-CoT} &  & 50.27  & 45.24   & N/A & \textbf{55.56}  & 27.84   & N/A  & \underline{48.99} & 45.58   \\
    {+ TCogMap} & &  \textbf{60.05} & \textbf{55.29} & N/A & 33.33 & \underline{37.65} & N/A & \textbf{62.91} & \underline{49.85} \\ \midrule
    \rowcolor{GroupBlue} \textit{InternVL3-8B} & Gen & \underline{37.77}	 & \textbf{56.48}& 	 N/A &            	\underline{44.44} &	31.86	& N/A  & 45.64	& 43.24 \\
    {+ Scene-CoT} &   & \textbf{55.62}  & 48.02    & N/A  & \textbf{55.56} &   \underline{33.57} & N/A  & 
\underline{50.34}  & \textbf{48.62}  \\
    {+ TCogMap} & & 37.23 & \underline{53.31}	& N/A  & \underline{44.44} & \textbf{39.98} & N/A  & \textbf{64.10}	& \underline{47.81} \\ 
    \rowcolor{GroupBlue}\textit{InternVL3-14B} & Gen &  40.58 & 43.92 & N/A             	& \underline{44.44}& 	\underline{43.53}	 & N/A& 43.90 & 43.27  \\       
    {+ Scene-CoT} &  & \textbf{53.17} & \underline{45.11}   & N/A  & \textbf{61.11}  & 31.76   & NA   & \underline{51.64}  & \textbf{48.56} \\
    {+ TCogMap} & &  \underline{46.47} & \textbf{48.41} & N/A & 38.89 & \textbf{45.87} & N/A & \textbf{59.00} & \underline{47.73} \\  
    \rowcolor{GroupBlue} \textit{InternVL3-38B} & Gen & 43.93 & \underline{48.28} & N/A & \underline{61.11} & \textbf{44.84} & N/A & \textbf{68.27} & \underline{53.29}  \\
    {+ Scene-CoT} &  & \textbf{55.07}  & 46.96   & N/A  & \underline{61.11}  &   33.42 &  N/A  & 53.77   & 50.07   \\
    {+ TCogMap} & & \underline{48.64} & \textbf{54.89} & N/A & \textbf{66.67} & \underline{39.23} & N/A & \underline{66.43} & \textbf{55.17} \\ \midrule \midrule
      \rowcolor{GroupBeige} \textit{RoboTron} & Drv &  23.73 & \underline{40.34} & N/A & \underline{33.33} & \underline{31.95} & N/A & 28.92 & 31.65  \\     
    {+ Scene-CoT} &  & \textbf{41.85} & 37.30  & N/A  & \textbf{55.56}    & 24.68    & N/A   & \underline{34.79}    & \textbf{38.84}  \\
    {+ TCogMap} & & \underline{24.37}	& \textbf{42.33}	 & N/A	& \underline{33.33}	& \textbf{35.21}	& N/A & \textbf{40.58} & \underline{35.16} \\ \midrule 
    \rowcolor{GroupBeige} \textit{
Cosmos-Reason} & Drv & 28.35	& 32.80	 & N/A & 44.44	& 16.18 & N/A & 16.06 & 27.57  \\
    {+ Scene-CoT} &   & \textbf{38.77} & \underline{34.13}  & N/A & \textbf{61.11} & \underline{24.10}  & N/A  & \textbf{43.72}  & \textbf{40.37}  \\
    {+ TCogMap} & & \underline{28.53} & \textbf{44.05} &	 N/A & \underline{50.00}	& \textbf{33.46} & N/A & \underline{23.97} & \underline{36.00} \\
        \bottomrule
    \end{tabular}
\label{tbl:non_ego_q_main_results}
\end{table*}
\renewcommand{\arraystretch}{1.0}

\subsection{Additional Experiments on CoT Descriptions for Scene-CoT}
In the main body of the paper, details of the four-step CoT reasoning process were described, providing motivation for dividing the problem into sub-tasks that focus on general scene descriptions, ego vehicle motion, motion of nearby vehicles and summarizing/formatting. However, one may question the following: ``Why was this four step process selected?'' and ``How sensitive are the benchmark results to the specific CoT prompts that are used?'' These questions are answered with this ablation study. 

Table~\ref{cot_ablation} compares different styles of descriptions generated during CoT reasoning for Scene-CoT. Three configurations were considered. With the Scene Description configuration, only the first step from the four-step CoT reasoning process was used to describe the video segments. In other words, for each video segment, only a high-level scene description was used for the LLM-based QA. This scenario can essentially be thought of as a CoT-free approach, where a general video description or caption is produced. In the second configuration, CoT Output Summary, only the final step of CoT reasoning was used for the segment descriptions during QA. The output from Step 4 is a formatted summary derived from the first three steps, so it is expected to be relatively complete. Finally, the ``Scene Description + CoT Output Summary'' configuration concatenates the output of Step 1 and Step 4 to form the segment descriptions. This is the configuration used for the results reported in the main body of the paper.

The average accuracies for these three segment description methods are shown in Table~\ref{cot_ablation}. The descriptions that consist of scene descriptions alone are relatively lower at 46.26\%. General scene descriptions are likely too impoverished to properly answer the breadth of questions considered in TAD. General scene descriptions tend to focus on the overall environment along with a few prominent scene objects. However, the question set of TAD focuses on many aspects of the scene that may not appear in a general description: vehicle actions, less prominent vehicles in the background, and ego vehicle actions. Using the final output from the CoT procedure shows a significant improvement, which is derived from a richer summary description that includes details of both the ego vehicle's and other scene vehicles' motions. However, best results are achieved when both the scene description and final CoT output are concatenated. This trend indicates that explicit scene-level context complements the structured reasoning provided by CoT, yielding more comprehensive representations for downstream question answering.
\begin{table}[tb!]
    \centering
    \caption{Effect of using different sections of the CoT outputs to form the segment descriptions.}
    \resizebox{\columnwidth}{!}{
    \begin{tabular}{l|c}
        \toprule
        \textbf{Description Style} 
        & \textbf{Average Accuracy}  \\
        \midrule
        {Scene Description} & 46.26  \\
        {CoT Output Summary} & 51.13  \\
        {Scene Description + CoT Output Summary} & 51.92 \\        
        \bottomrule
    \end{tabular}
    }
\label{cot_ablation}
\end{table}


\subsection{Object detections as context}
This ablation considers the possibility of leveraging knowledge from an ``external expert'' detector agent to improve the VLM performance in the proposed methods.  In this case, ablation experiments were completed on a modified version of the Scene-CoT pipeline to explore the effect on reasoning trace quality when information from an object detector is provided.

As shown in \cref{fig:Cap4Time}, the VLM CoT Reasoning module is used to generate a descriptive reasoning trace for each video segment, $s_j$, which are subsequently used by an LLM to perform question answering. In the Scene-CoT approach, for a specific video segment, four keyframes are provided to the VLM to perform the four steps of the CoT reasoning.  In this ablation, one additional pre-processing step is added prior to VLM CoT Reasoning.  Specifically, the Ultralytics YOLO11x \cite{yolo11_ultralytics} object detector is first run on each of the keyframes in a segment. The output of the detector is then post-processed to provide a textual summary denoting the count of the relevant vehicle categories that are visible within that keyframe. The relevant object categories are defined as those that overlap with the vehicle categories used in TAD, which are: bicycle, car, motorcycle, bus, and truck.  As an explicit example, the post-processed detector output for one keyframe has the format `Frame1: 5 cars, 2 bicycles'. This object detection summary string is computed for each of the keyframes in a video segment and then provided as additional context to the first step of the CoT Reasoning module that computes the summary Scene Description. The rationale behind this design is that the expert object detector can provide a prior to the VLM that can resolve ambiguities in challenging situations when the VLM is less confident about the presence of certain objects.

This ablation study was performed using four VLMs, Qwen2.5-VL-7B/32B and InternVL3-8B/14B, the results of which are shown in \cref{yolo_ablation}. It appears there may be some promise in adding external knowledge via an expert model, since the addition of the YOLO11x detection summary improved average accuracy across all four tested models. However, the gains are quite modest, with Qwen2.5-VL-32B seeing the biggest improvement of 0.78 and an average improvement of only 0.53 across the four models.  Overall, this result suggests that although there is some promise, the inclusion of an additional detector agent may not be worth the extra system complexity.

\begin{table}[h]
    \centering   
    \caption{Effect of adding object detector context as input to the VLM CoT Reasoning module. Numbers denote the average accuracy across all seven tasks in the TAD benchmark.}
    \resizebox{\columnwidth}{!}{
    \begin{tabular}{l|c|c|c}
        \toprule
        \textbf{VLM Model} 
        & \textbf{Scene-CoT}  
        & \textbf{Scene-CoT + YOLO11x} 
        & \textbf{Difference} \\
        \midrule
        \textit{Qwen2.5-VL-7B} & 48.71 & 49.17 & 0.46  \\
        \textit{Qwen2.5-VL-32B} & 48.78 & 49.56 & 0.78 \\        
        \textit{InternVL3-8B} & 51.92 & 52.68 & 0.76 \\
        \textit{InternVL3-14B } & 51.81 & 51.93 & 0.12 \\    
        \midrule
        \textbf{Average} & 50.31 & 50.84 & 0.53 \\  
        \bottomrule
    \end{tabular}
    }
\label{yolo_ablation}
\end{table}

\subsection{Inference Time Analysis}
{
While inference time was not a primary focus of our evaluation, we further analyzed the inference time for completeness. To measure the inference time, roughly 200 questions across three different question types} (\textit{Ego Action Duration}, \textit{Relative Temporal Localization}, and \textit{Temporal Ordering}) were used, the results for which are shown in \cref{table_inference_time} 

Four configurations were considered: the two baselines (with and without additional ego pose context) and the two proposed solutions (Scene-CoT and TCogMap). As the table shows, the baselines and TCogMap have quite comparable inference times. This result is due to the fact that the image tokens for all three methods are identical but the textual tokens are slightly different, depending on the specific contextual information that is added by the method. TCogMap has the additional overhead of computing the ego vehicle temporal cognitive map and classifying the ego vehicle actions, but these processing steps have a negligible effect on latency. 

For Scene-CoT, on the other hand, a much higher inference time is observed. This large inference time is primarily due to the increased number of VLM inference calls, since CoT descriptions are provided for each segment of the video. Specifically, for the experiments in this paper, the videos are divided into ten segments, each of which is described using four CoT processing steps, yielding a total of 40 VLM inference calls. This design results in an inference time of 42 seconds per question to generate the segment descriptions. In comparison, the actual LLM question-answering inference time is rather minor, at under 5 seconds per question.

Although this run-time of Scene-CoT can be viewed as a disadvantage, there is significant ongoing research on VLM inference optimization, including quantization (e.g., \cite{frantar2022gptq,tseng2024quip,egiazarian2024extreme,gholami2025casp}) and pruning \cite{chen2024image,lin2025boosting,alvar2025divprune}, that can provide significant speed improvements with limited accuracy losses. Moreover, the experiments in this paper were performed in a single inference setting, but batch inference could be used to parallelize the description generation across the segments, resulting in significant inference speed improvements (i.e., up to 10$\times$, in the case of the experiments presented in this paper). Thus, there is significant scope to reduce the runtime of Scene-CoT.
\begin{table}[tb!]
    \centering
    \caption{Average inference time for the baselines and proposed solutions.}    
    \begin{tabular}{l|c}
        \toprule
        \textbf{Method} 
        & \textbf{Inference time (s)}  \\
        \midrule
        \rowcolor{GroupBlue} \textit{InternVL3-8B} & 2.20  \\
        \rowcolor{GroupBlue} \textit{InternVL3-8B + EP} & 2.72  \\
        {+ Scene-CoT} &  47.10 \\
        {+ TCogMap} & 2.21  \\
        \bottomrule
    \end{tabular}
\label{table_inference_time}
\end{table}

\subsection{Effect of LLM on Scene-CoT}
In the main body of the paper, the results for Scene-CoT were generated using Qwen2.5-14B-Instruct-1M as the LLM for question answering. This LLM was selected through an empirical evaluation process.  This procedure, as well as the effects of utilizing different LLMs, are described next.

In~\cref{llm_ablation}, the effect of changing the question answering LLM for Scene-CoT is shown. A clear accuracy improvement of 2.21\% was observed when moving from the 7B Qwen2.5-7B-Instruct to the comparable 14B model, Qwen2.5-14B-Instruct. This observation motivated the selection of a 14B LLM that can offer greater reasoning power. Among the 14B models evaluated, Qwen2.5-14B-Instruct-1M achieves the highest average accuracy of 51.92\%. The comparison between the standard and ``1M'' versions of the 14B model indicates that extended context capacity and stronger instruction tuning improve reasoning over temporally grounded inputs. Additionally, Qwen3-14B with its ``thinking'' mode enabled achieves comparable accuracy to Qwen2.5-14B-Instruct-1M; however, Qwen2.5-14B-Instruct-1M was ultimately selected due to its higher throughput. 

Overall, the trends indicate that there is a 3-4\% improvement when moving to the larger 14B LLMs and that amongst the 14B models, the accuracies are fairly comparable.
\begin{table}[h]
    \centering    
    \caption{Effect of using different LLMs for question answering in Scene-CoT. Average accuracy is computed over all seven tasks.}
    \begin{tabular}{l|c}
        \toprule
        \textbf{LLM QA Model} 
        & \textbf{Average Accuracy}  \\
        \midrule
        \textit{Qwen2.5-7B-Instruct} & 47.87  \\
        \textit{Qwen2.5-14B-Instruct} & 50.08 \\        
        \textit{Qwen2.5-14B-Instruct-1M} & \textbf{51.92} \\
        \textit{Qwen3-14B} (thinking enabled) & \underline{51.05} \\    
        \bottomrule
    \end{tabular}
\label{llm_ablation}
\end{table}

\section{Data Annotation Interface}
\begin{figure*}[tb!]
 \centering
  \includegraphics[width=.95\textwidth]{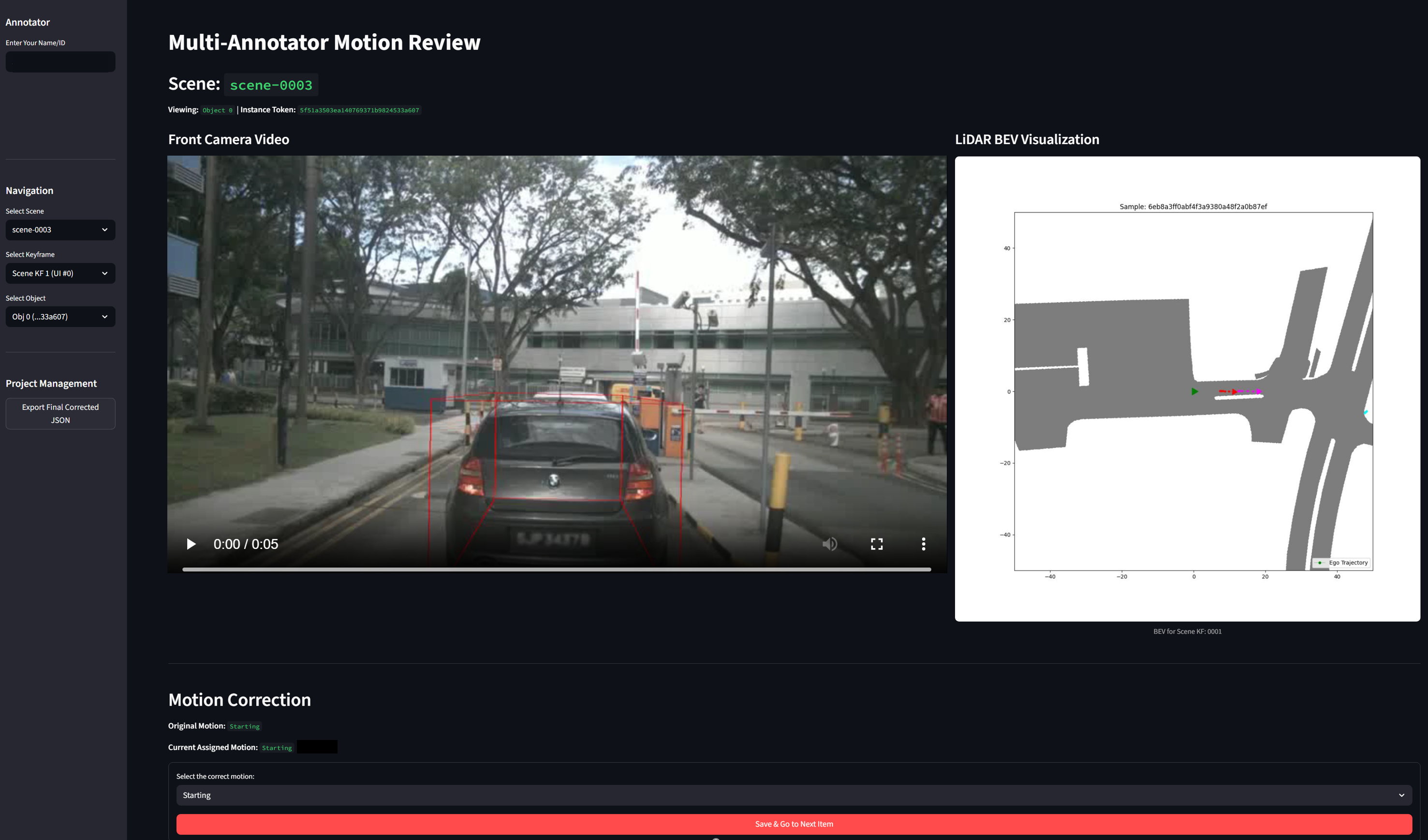}
 \caption{Data annotation pipeline.}
 \label{fig:annotation_demo}
 \end{figure*}

\cref{fig:annotation_demo} illustrates the annotation interface used to collect vehicle-action labels for TAD. Since labeling the actions of specific vehicles, including the ego vehicle, requires a level of granularity not supported by existing annotation tools, a custom interface is implemented. Each video segment is displayed alongside its corresponding bird’s-eye-view (BEV) visualization, which shows the ego trajectory and the motion of nearby vehicles. Annotators view both modalities jointly to assess the dominant behavior of each target vehicle within the segment. A drop-down menu is provided for selecting the appropriate motion label for the target object, and the final annotation is saved through the interface. This setup ensures consistent access to both appearance and trajectory cues, facilitating reliable segment-level action labeling.

\section{Qualitative Examples}
\begin{figure*}[tb!]
 \centering
  \includegraphics[width=1.0\textwidth]{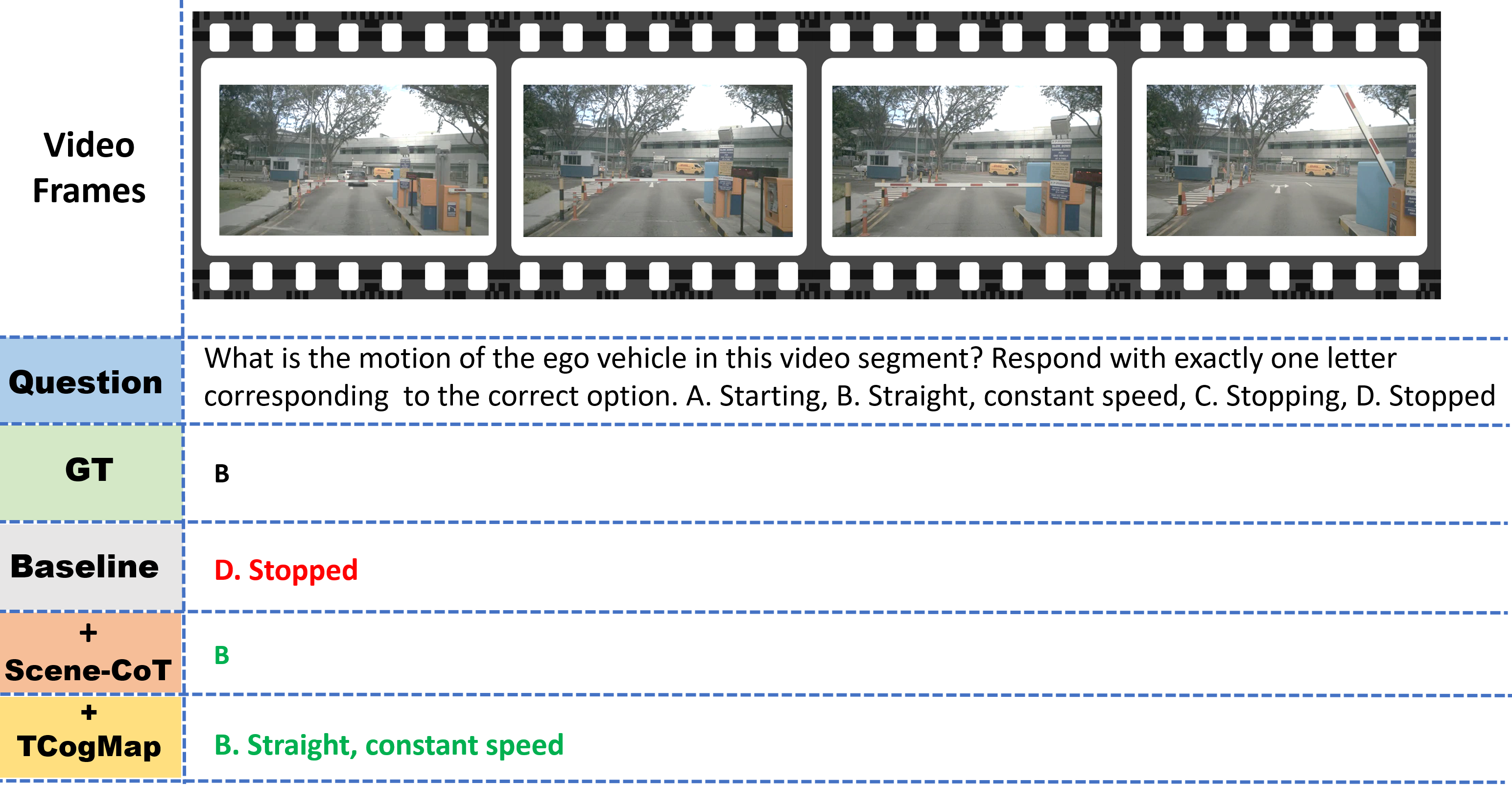}
 \caption{Qualitative example for \textit{Multiple Choice Action Recognition} task using the baseline (InternVL3-8B) and the proposed methods. \textbf{\textcolor{ForestGreen}{Green}} text indicates a correct answer, \textbf{\textcolor{red}{red}} denotes an incorrect answer.}
 \label{fig:qualitative_mc_action}
 \end{figure*}

\begin{figure*}[tb!]
 \centering
  \includegraphics[width=1.0\textwidth]{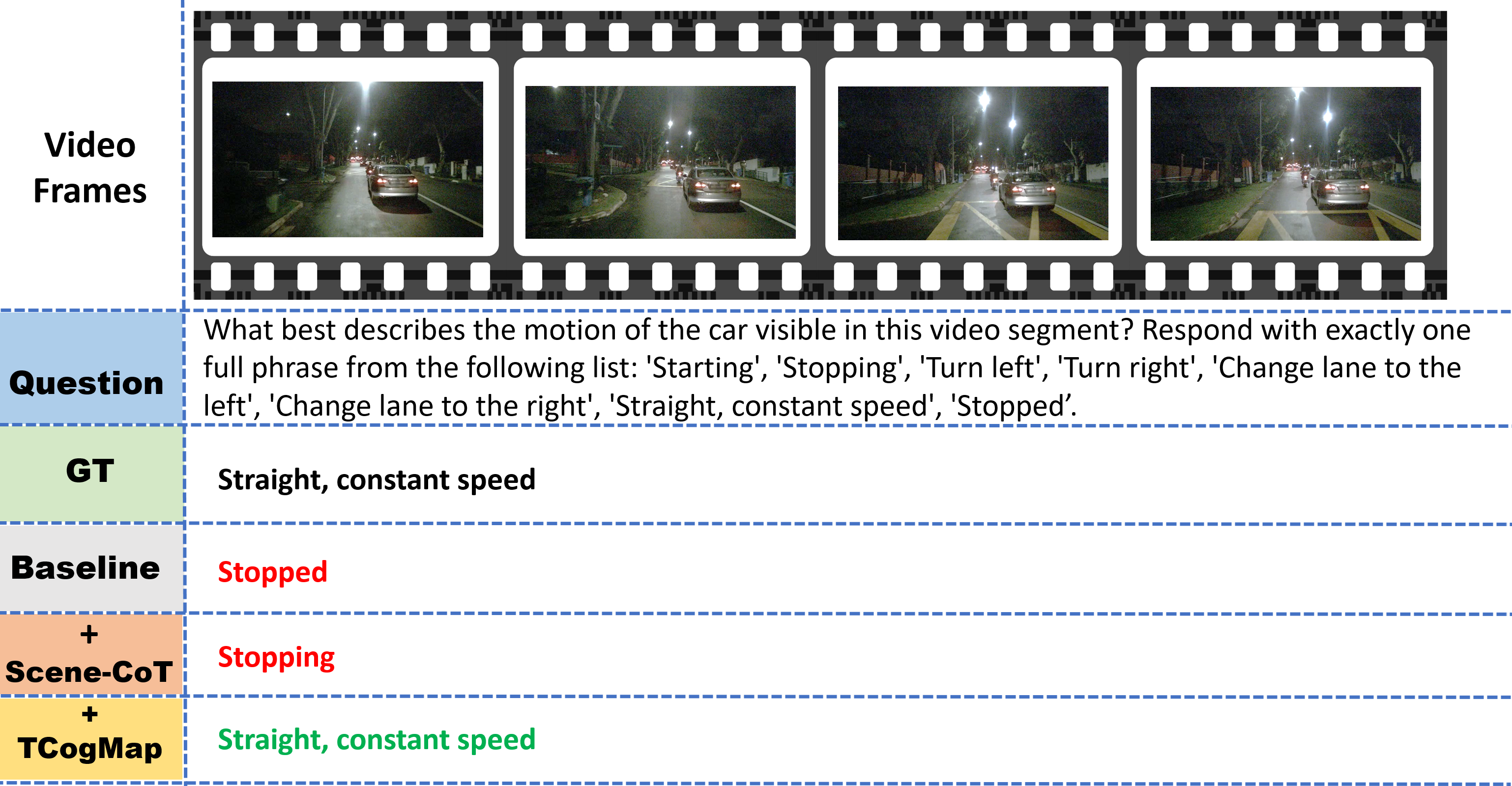}
 \caption{Qualitative example for \textit{Exact Answer Action Recognition} task using the baseline (InternVL3-8B) and proposed methods. \textbf{\textcolor{ForestGreen}{Green}} text indicates a correct answer, \textbf{\textcolor{red}{red}} denotes an incorrect answer.}
 \label{fig:qualitative_exact_action}
 \end{figure*}

\begin{figure*}[tb!]
 \centering
  \includegraphics[width=1.0\textwidth]{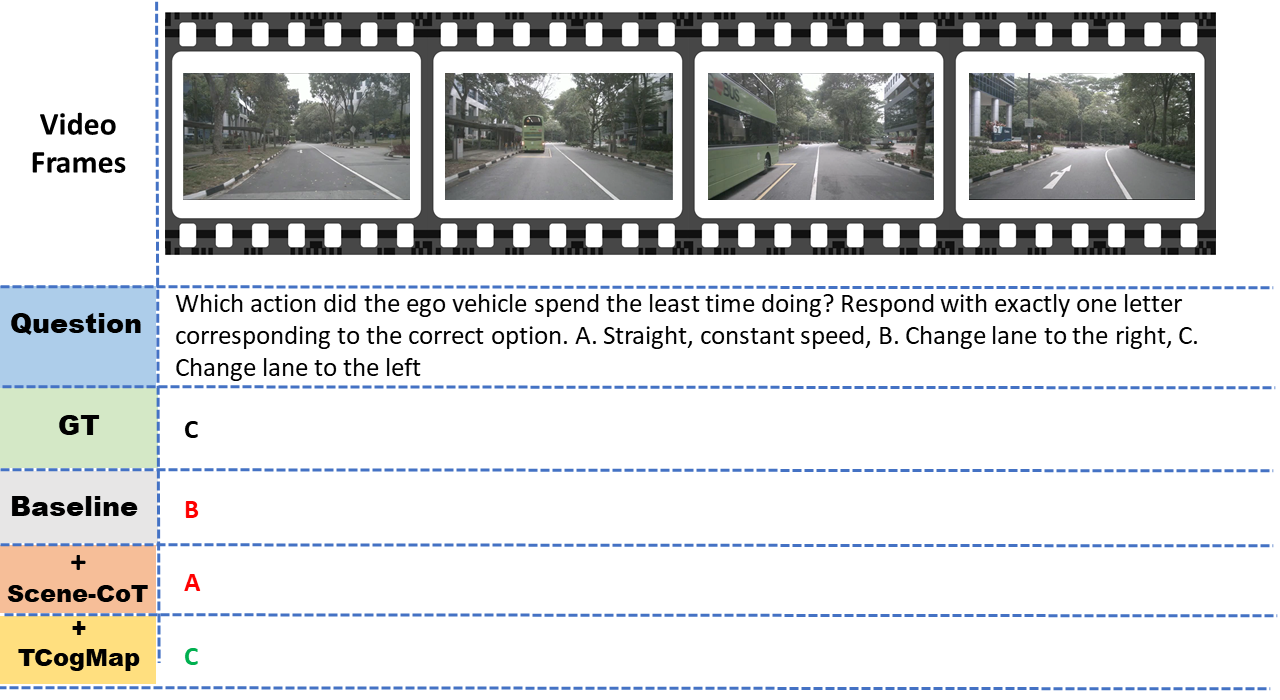}
 \caption{Qualitative example for \textit{Action Duration} task using the baseline (InternVL3-8B) and proposed methods. \textbf{\textcolor{ForestGreen}{Green}} text indicates a correct answer, \textbf{\textcolor{red}{red}} denotes an incorrect answer.}
 \label{fig:qualitative_action_duration}
 \end{figure*}

\begin{figure*}[hptb!]
 \centering
  \includegraphics[width=1.0\textwidth]{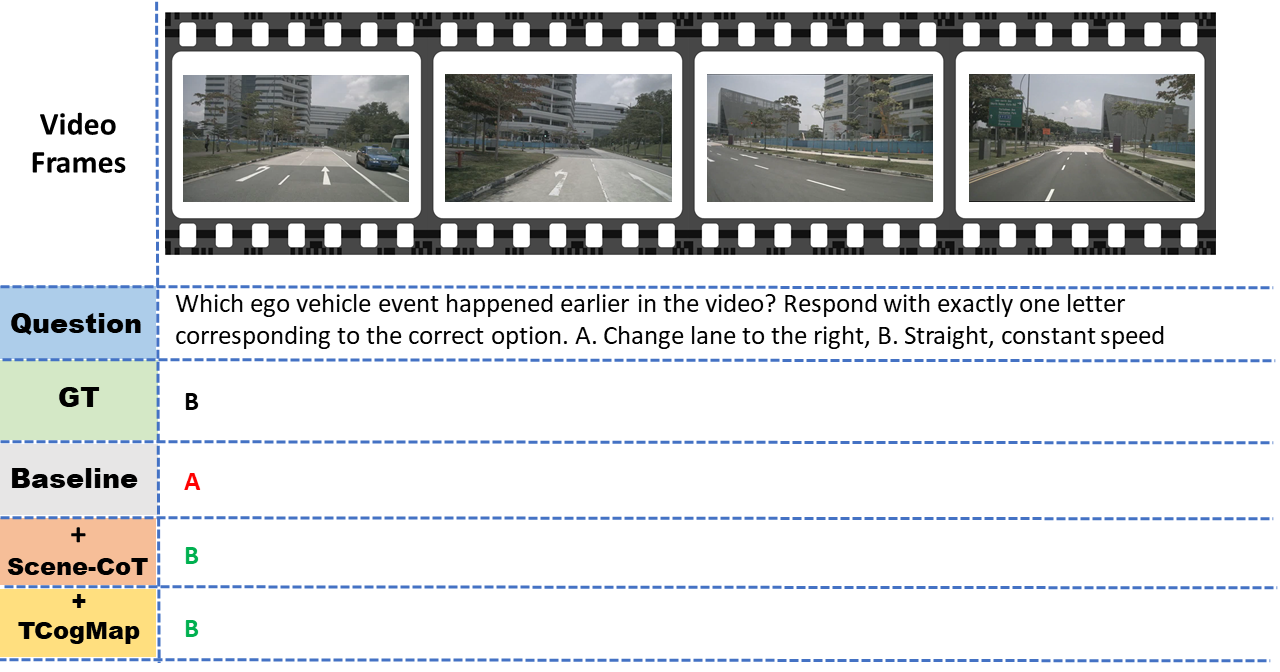}
 \caption{Qualitative example for \textit{Relative Temporal Action Localization} task using the baseline (InternVL3-8B) and proposed methods. \textbf{\textcolor{ForestGreen}{Green}} text indicates a correct answer, \textbf{\textcolor{red}{red}} denotes an incorrect answer.}
 \label{fig:qualitative_relative_temporal}
 \end{figure*}

Figures~\ref{fig:qualitative_mc_action}–\ref{fig:qualitative_relative_temporal} present qualitative examples corresponding to Tasks \textit{Multiple Choice Action Recognition}, \textit{Exact Answer Action Recognition}, \textit{Action Duration}, and \textit{Relative Temporal Action Localization}, respectively, complementing the quantitative results in the main table of the paper. Each example includes representative frames from the relevant segment or full scene, the question posed, the ground-truth answer, and the model (i.e, InternVL3-8B) outputs under different inference settings. The baseline corresponds to the model receiving only the frames and the question. The “+Scene-CoT” setting applies the proposed Scene-CoT reasoning procedure, while “+TCogMap” shows the outputs produced using the proposed TCogMap. Across tasks, the qualitative results show that applying Scene-CoT or TCogMap improves the model’s predictions, often correcting cases where the baseline fails, underscoring the value of both methods for strengthening temporal and scene-level reasoning.

In more detail, in \cref{fig:qualitative_mc_action}, the ego vehicle goes straight toward a barrier, which then lifts to allow the ego vehicle to pass. The baseline selects an incorrect answer, whereas both Scene-CoT and TCogMap produce the correct output. In \cref{fig:qualitative_exact_action}, the query concerns the motion of the car visible in the frames. Because the scene occurs at night and the rear lights of the vehicle are illuminated red, both the baseline and Scene-CoT incorrectly predict that the car is ``Stopped.'' In contrast, TCogMap leverages the ego trajectory and scene dynamics to correctly identify the action of the vehicle ahead as ``Straight, constant speed''. In \cref{fig:qualitative_action_duration}, the video shows the ego vehicle first approaching a parked bus. The ego vehicle performs a long, gradual lane change to the right to avoid the bus, followed by a quick lane change back to the left. The baseline and Scene-CoT both predict incorrect answers, whereas TCogMap is able to differentiate between the left and right lane changes to provide the correct answer of ``Change lane to the left''. In \cref{fig:qualitative_relative_temporal}, the video first shows the ego vehicle driving straight down a city road. Thereafter, the ego vehicle performs a series of short, consecutive atomic actions: a lane change to the left, followed by a left hand turn around a corner, and finally a lane change to the right. The baseline incorrectly selects ``Change lane to the right'' as its answer. In contrast, Scene-CoT and TCogMap are both able to correctly assess that the ego vehicle drives straight at the beginning of the video, while the lane change to the right occurs near the end, thus allowing them both to provide the correct answer of ``B''. 

\begin{figure*}[tb!]
 \centering
  \includegraphics[height=0.85\textheight,keepaspectratio]{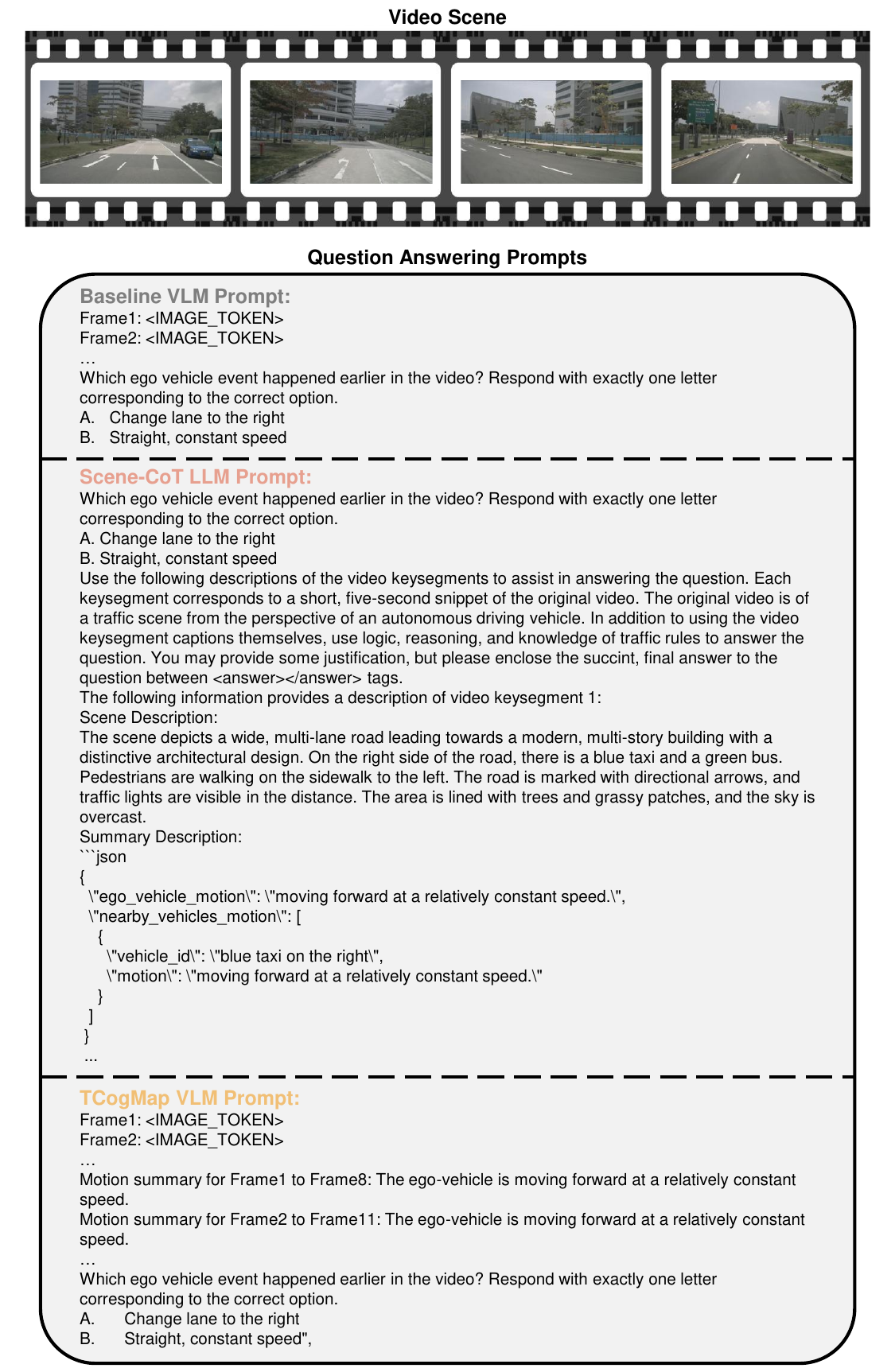}
\vspace{-4pt}
 \caption{Prompts used for question answering. Note that for the baseline and TCogMap, these prompts are provided to the VLM as input, whereas for Scene-CoT, this prompt is given to the LLM}
 \label{fig:qa_prompt} 
 \end{figure*}

A sample of the exact prompts used during the question answering step for the baseline and two proposed approaches is shown in \cref{fig:qa_prompt}, depicting the same video scene that appears in \cref{fig:qualitative_relative_temporal}.  Note that for the baseline and TCogMap approaches, question answering is performed via a VLM, whereas for Scene-CoT, the final step of answering the question is completed by an LLM.

As \cref{fig:qa_prompt} shows, the prompt for the baseline method is straightforward, consisting merely of the frame labels and corresponding image token tags, as well as the question text itself. TCogMap extends the baseline prompt by adding a motion summary, derived from the temporal cognitive map, for each segment of the video.  As shown quantitatively in the main body of the paper, this additional context significantly improves the average accuracy on TAD for all of the VLMs that were considered. For Scene-CoT, since the prompt is being used with an LLM, the reference to frames and image tokens are removed.  Instead, the prompt consists of the question itself along with textual descriptions for each video segment. These descriptions were formed using the proposed CoT procedure and include a scene description as well as a JSON-style summary. As shown in the main body of the paper, Scene-CoT, with its CoT reasoning process and LLM-based question answering, generally improves accuracies on the TAD benchmark for smaller models.

\clearpage


\end{document}